%% file: main.tex
\documentclass[lettersize,journal]{IEEEtran}
\usepackage{amsmath,amsfonts}
\usepackage{algorithmic}
\usepackage{algorithm}
\usepackage{array}
\usepackage[caption=false,font=footnotesize,labelfont=rm,textfont=rm]{subfig}
\usepackage{textcomp}
\usepackage{stfloats}
\usepackage{url}
\usepackage{verbatim}
\usepackage{graphicx}
\usepackage{cite}
\usepackage{booktabs}   
\usepackage{multirow}   
\usepackage{array}      
\usepackage[colorlinks=true, allcolors=blue]{hyperref}
\usepackage{colortbl}
\usepackage{xcolor}
\usepackage{pifont}
\usepackage{tabularx}
\usepackage{float}
\usepackage{dashrule}

\begin{document}

\title{M\textsuperscript{2}IR: Proactive All-in-One Image Restoration via Mamba-style Modulation and Mixture-of-Experts}

\author{Shiwei Wang, Yongzhen Wang, Bingwen Hu, Liyan Zhang, Xiao-Ping Zhang, \IEEEmembership{Fellow,~IEEE}, and Mingqiang Wei, \IEEEmembership{Senior Member,~IEEE}
\thanks{Shiwei Wang and Liyan Zhang are with the School of Computer Science and Technology, Nanjing University of Aeronautics and Astronautics, Nanjing 210016, China {(e-mail: wangshiwei@nuaa.edu.cn; zhangliyan@nuaa.edu.cn).}}
\thanks{Yongzhen Wang and Bingwen Hu are with the School of Computer Science and Technology, Anhui University of Technology, Ma’anshan 243032, China (e-mail: wangyz@ahut.edu.cn; hu\_bingwen@ahut.edu.cn).}
\thanks{Xiao-Ping Zhang is with the Shenzhen Ubiquitous Data Enabling Key Laboratory, Tsinghua Shenzhen International Graduate School, Tsinghua University, Shenzhen 518055, China (e-mail: xpzhang@ieee.org).}
\thanks{Mingqiang Wei is with the School of Computer Science and Technology, Nanjing University of Aeronautics and Astronautics, Nanjing 210016, China, and also with the College of Artificial Intelligence, Taiyuan University of Technology, Taiyuan 030024, China (e-mail: mingqiang.wei@gmail.com).}
}

\markboth{Journal of \LaTeX\ Class Files,~Vol.~14, No.~8, August~2021}%
{Shell \MakeLowercase{\textit{et al.}}: A Sample Article Using IEEEtran.cls for IEEE Journals}


\maketitle
\input{sec/0_abstract}

\input{sec/1_introduction}
\input{sec/2_related_work}
\input{sec/3_method}
\input{sec/4_experiments}
\input{sec/5_conclusion}

\bibliographystyle{IEEEtran}
\bibliography{main}

\end{document}

%% file: sec/0_abstract.tex
\begin{abstract}
While Transformer-based architectures have dominated recent advances in all-in-one image restoration, they remain fundamentally reactive: propagating degradations rather than proactively suppressing them. In the absence of explicit suppression mechanisms, degraded signals interfere with feature learning, compelling the decoder to balance artifact removal and detail preservation, thereby increasing model complexity and limiting adaptability. To address these challenges, we propose M\textsuperscript{2}IR, a novel restoration framework that proactively regulates degradation propagation during the encoding stage and efficiently eliminates residual degradations during decoding. Specifically, the Mamba-Style Transformer (MST) block performs pixel-wise selective state modulation to mitigate degradations while preserving structural integrity. In parallel, the Adaptive Degradation Expert Collaboration (ADEC) module utilizes degradation-specific experts guided by a DA-CLIP-driven router and complemented by a shared expert to eliminate residual degradations through targeted and cooperative restoration. By integrating the MST block and ADEC module, M\textsuperscript{2}IR transitions from passive reaction to active degradation control, effectively harnessing learned representations to achieve superior generalization, enhanced adaptability, and refined recovery of fine-grained details across diverse all-in-one image restoration benchmarks. 
Our source codes are available at \href{https://github.com/Im34v/M2IR}{https://github.com/Im34v/M2IR}.
\end{abstract}

%% file: sec/1_introduction.tex
\section{Introduction}

Image restoration is a fundamental task in low-level vision that aims to recover high-quality images from degraded observations, with broad applications in medical imaging, remote sensing, and autonomous perception.
While deep neural networks have greatly advanced this field through end-to-end learning, most existing approaches remain task-specific, focusing on individual degradations such as image denoising \cite{zhang2017beyond,zhang2018ffdnet,shen2023adaptive}, dehazing \cite{cai2016dehazenet,li2017aod,dong2020fd,song2023vision}, or deraining \cite{yasarla2019uncertainty,fu2019lightweight,jiang2020multi,chen2023learning}.
Although effective within their respective domains, these models struggle to generalize to unseen or mixed degradations, motivating the development of all-in-one image restoration capable of addressing diverse degradation types within a unified model.

\begin{figure}[t]
    \centering
    \includegraphics[width=0.95\linewidth]{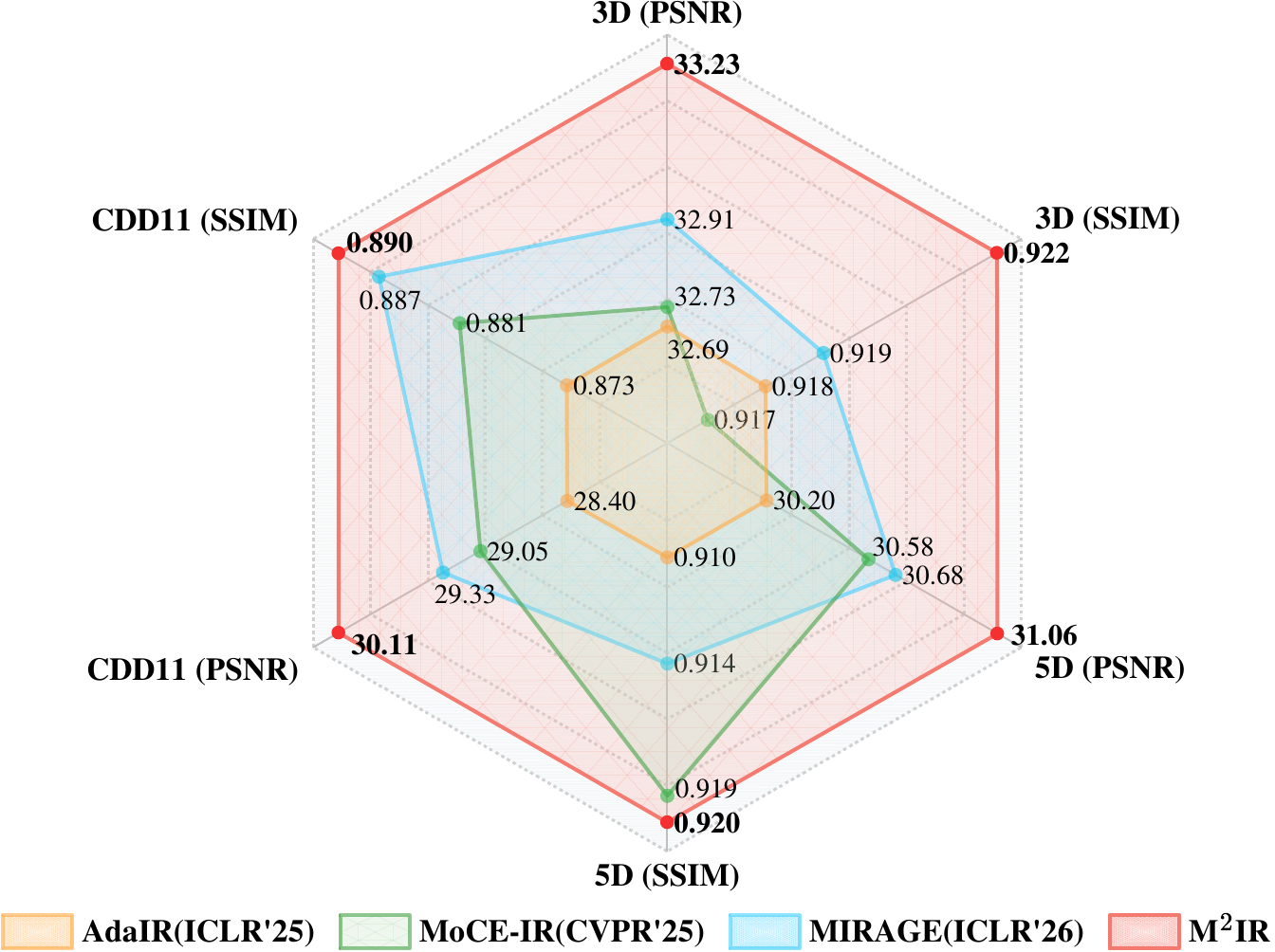}
    \caption{Performance comparison of recent all-in-one restoration methods under different settings. Here, 3D indicates three degradation types encompassing haze, rain, and noise, while 5D extends 3D by adding blur and low-light. The CDD11 dataset contains four fundamental degradation types, low-light, hazy, rainy, and snowy, as well as seven composite degradation types generated by combining these fundamental types. 
    }
    \label{fig: compare}
\end{figure}

Recent approaches \cite{potlapalli2023promptir, yu2024multi, cui2025adair, zamfir2025complexity, zhang2025perceive} have primarily focused on leveraging efficient Transformer architectures \cite{zamir2022restormer} with channel-wise attention as encoder layers, providing linear complexity and strong global context modeling. Despite their success, these models remain inherently passive in addressing degradation, lacking mechanisms for proactive suppression or regulation of degradation information. Conventional encoder designs indiscriminately encode both clean scene content and degradation patterns into a unified latent representation. As features propagate through successive layers, degradation signals become increasingly mixed with structural details, leading to entangled representations that obscure the underlying scene. Consequently, the decoder must devote additional computational effort to disentangle degradations from meaningful image content before reconstruction, thereby increasing decoding complexity and amplifying the risk of residual artifacts. 
 
To enhance decoder adaptability and representation capacity for challenging all-in-one restoration tasks, recent studies have introduced prompt learning \cite{potlapalli2023promptir, kong2024towards} and feature modulation \cite{cui2025adair} strategies, yielding notable improvements. However, these methods remain reactive in nature, mitigating degradation after feature contamination instead of proactively preventing its propagation during encoding. This inherent passivity compromises representational clarity and makes precise restoration substantially more challenging. Furthermore, existing approaches \cite{potlapalli2023promptir, kong2024towards, cui2025adair} typically adopt a single unified decoder for feature reconstruction, which often underutilizes features and struggles with spatially varying degradations. The Mixture-of-Experts (MoE) framework \cite{yu2024multi, zhang2024efficient} has recently shown strong potential in image restoration by enabling dynamic sparse routing and parallel experts. This design adaptively selects task-specific weights on demand, enabling efficient and robust end-to-end restoration across multiple degradation types. However, while specialized experts are effective for particular degradations, many degradations share common patterns. If the routing mechanism fails to manage these overlaps properly, certain experts may be repeatedly activated for degradations beyond their specialization, leading to expert load imbalance and degraded performance. 

To address these challenges, we propose M\textsuperscript{2}IR, a degradation-aware restoration framework that transforms passive compensation into proactive degradation governance. Instead of allowing degradation to propagate unchecked and compensating for it afterward, M\textsuperscript{2}IR enforces proactive suppression during the encoding phase to constrain the spread of degraded information. Nevertheless, such suppression cannot completely eliminate degradation, i.e., residual noise inevitably persists within latent representations. 
To effectively refine these residuals, we extend the conventional Mixture-of-Experts (MoE) paradigm by integrating both specialized and shared experts, enabling adaptive collaboration across diverse degradation patterns. Specifically, we revisit Transformer-based architectures \cite{zamir2022restormer} through the lens of degradation inhibition and introduce the \textbf{Mamba-Style Transformer (MST)}, which performs input-dependent dynamic modulation via pixel-wise selective state modulation. This mechanism guides the model to emphasize structural and semantic cues while attenuating the propagation of degradation-related signals. Building upon this foundation, we further develop the \textbf{Adaptive Degradation Expert Collaboration (ADEC)} module. ADEC incorporates a DA-CLIP–guided router \cite{luo2023controlling} that leverages degradation-aware contextual priors to accurately identify degradation types and balance expert activation, thereby mitigating expert overuse and fostering cooperative restoration. Through this collaborative design, M\textsuperscript{2}IR achieves a synergistic balance between degradation-specific precision and general restoration robustness, establishing a strong foundation for generalization across diverse degradation scenarios.
Extensive experiments validate the effectiveness of M\textsuperscript{2}IR, demonstrating state-of-the-art performance across multiple all-in-one restoration benchmarks, as illustrated in Fig.~\ref{fig: compare}.
In summary, the main contributions of this work are as follows:
\begin{itemize}
    \item We reformulate all-in-one image restoration from a passive compensation paradigm into a proactive degradation governance framework by introducing M\textsuperscript{2}IR, which suppresses degradation propagation through the Mamba-Style Transformer (MST) and achieves adaptive decoding via the Adaptive Degradation Expert Collaboration (ADEC) module.
    \item We develop the Mamba-Style Transformer (MST) from the perspective of degradation inhibition, leveraging pixel-wise selective state modulation to proactively suppress degradation-related signals while maintaining structural integrity throughout the encoding stage.
    \item We propose the Adaptive Degradation Expert Collaboration (ADEC) module, which employs a DA-CLIP–guided router to coordinate specialized and shared experts, effectively eliminating residual degradations, balancing expert utilization, and enhancing restoration robustness across diverse degradation scenarios.
\end{itemize}

\begin{figure*}[t]
    \centering
    \includegraphics[width=1\linewidth]{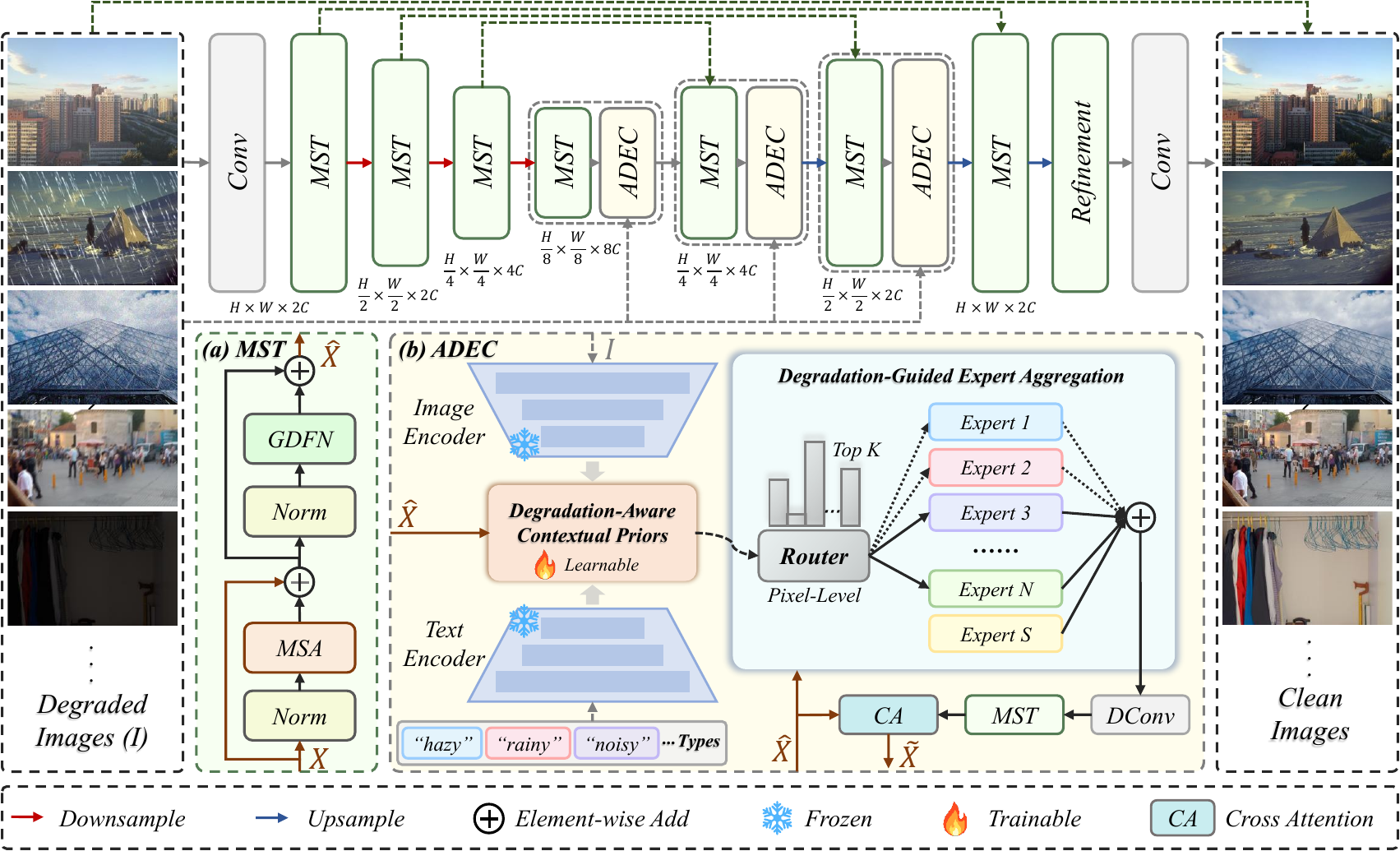}
    \caption{Overview of the proposed M\textsuperscript{2}IR. 
    (a) Mamba-Style Transformer (MST): Combining the strengths of Mamba and Transformer, it achieves proactive suppression or regulation of degradation information. 
    (b) Adaptive Degradation Expert Collaboration (ADEC): It utilizes DA-CLIP's image encoder and text encoder to obtain degradation-aware contextual priors, further deriving routing weights, and then aggregates the results from the selected $K$ most relevant experts and a shared expert to eliminate residual degradations effectively. 
}
    \label{fig: M2IR_Pipeline}
\end{figure*}

%% file: sec/2_related_work.tex
\section{Related Work}

\subsection{All-in-One Image Restoration}

Vision Transformers (ViTs) \cite{dosovitskiy2020image} have greatly advanced image restoration by effectively modeling long-range dependencies. Representative works such as Restormer \cite{zamir2022restormer} have demonstrated the potential of Transformer-based architectures across diverse restoration tasks, inspiring the development of unified models for handling multiple degradations. 
Early all-in-one approaches, including AirNet \cite{li2022all} and IDR \cite{zhang2023ingredient}, integrate degradation recognition with restoration through contrastive learning or progressive disentanglement.
More recent efforts have introduced prompt learning and adaptive modulation to enhance multi-task restoration performance. For instance, PromptIR \cite{potlapalli2023promptir}  employs dynamic prompts, while AutoDIR \cite{jiang2023autodir} and DA-CLIP \cite{luo2023controlling} exploit pre-trained models and contrastive learning to handle open-vocabulary or mixed degradations. MiOIR \cite{kong2024towards} and AdaIR \cite{cui2025adair} further refine multi-task learning and degradation adaptation, improving generalization for complex scenarios.
Despite these advances, most existing frameworks remain reactive, passively propagating degradation features without explicit regulation. In contrast, our M\textsuperscript{2}IR introduces proactive degradation governance, explicitly suppressing degraded signals during encoding and adaptively refining residuals during decoding, thereby achieving more robust and generalizable image restoration.

\subsection{Mamba in Vision}

Mamba \cite{mamba}, a selective State Space Model (SSM), has recently shown strong potential in vision tasks following its success in language modeling. By dynamically controlling information propagation through input-dependent gating, Mamba effectively captures long-range dependencies with linear computational complexity. Extensions such as VMamba \cite{liu2024vmamba} and Vision Mamba \cite{zhu2024vision} adopt this paradigm to visual domains via cross-scanning and bidirectional mechanisms. However, these variants still flatten 2D structures into 1D sequences, which undermines spatial consistency. Further analysis by MILA \cite{han2024demystify} reveals that the forget gate and block composition are crucial for selective state propagation, emphasizing Mamba’s strength in efficient context modeling. These insights inspire our Mamba-Style Transformer, which reformulates state modulation to suppress degradation while preserving spatial structure.

\subsection{Mixture-of-Experts for Image Restoration}

The Mixture-of-Experts (MoE) framework decomposes representation learning into multiple experts, guided by a routing mechanism that adaptively activates task-relevant modules. Owing to its dynamic specialization and scalability, MoE is inherently well-suited for all-in-one image restoration. Recent studies, such as MEASNet \cite{yu2024multi} and MoFME \cite{zhang2024efficient}, have enhanced expert routing through task- and feature-aware strategies, enabling adaptive computation across diverse inputs. However, these approaches often lack coordination between specialized and shared experts, resulting in imbalanced expert utilization and limited collaborative refinement. To overcome these limitations, we propose the Adaptive Degradation Expert Collaboration (ADEC) module, which introduces degradation-aware contextual routing to promote balanced expert activation and cooperative restoration. 

%% file: sec/3_method.tex
\section{Method}
In this section, we first describe the overall architecture of M\textsuperscript{2}IR. We then detail its two core components: the Mamba-Style Transformer (MST) block and the Adaptive Degradation Expert Collaboration (ADEC) module. 

\subsection{Overall Framework}

As illustrated in Fig. \ref{fig: M2IR_Pipeline}, M\textsuperscript{2}IR adopts a U-shaped architecture to restore degraded inputs into clean images in an end-to-end manner. 
Given a degraded input $I \in \mathbb{R}^{H\times W\times 3}$, shallow features $F_0 \in \mathbb{R}^{H\times W\times C}$ are first extracted using a $3 \times 3$ convolution, where $H \times W$ denotes spatial resolution and $C$ the channel dimension. 
The features are then processed by a 4-stage encoder-decoder network, with each stage containing multiple MST blocks. 
To enhance adaptability under diverse degradation scenarios, we introduce the Adaptive Degradation Expert Collaboration (ADEC) module between successive decoder stages. 
ADEC leverages degradation-aware priors from pre-trained DA-CLIP's image encoder and text encoder to generate the routing weights, which are considered as confidence scores for each expert. 
For each pixel, the top $K$ experts with the highest scores are selected to process the degradation with a shared expert, and their outputs are fused adaptively. 
This design tightly integrates task-specific expert allocation with shared experts, effectively tackling diverse degradations while enhancing the network’s flexibility and boosting restoration performance within a unified framework. 

\subsection{Mamba-Style Transformer}

\begin{figure}[t]
    \centering
    \includegraphics[width=1\linewidth]{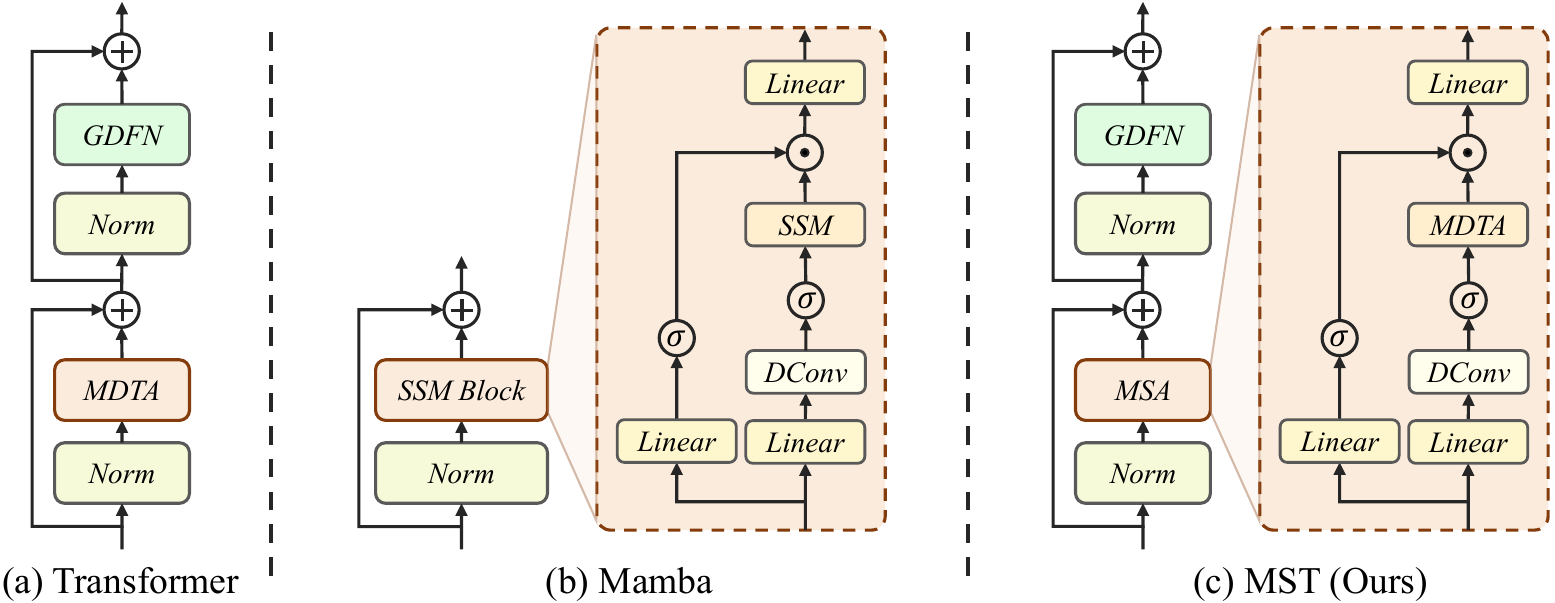}
    \caption{Macro architecture illustrations of (a) Transformer \cite{zamir2022restormer} with MDTA and DGFN, (b) Mamba \cite{mamba} with SSM, and (c) the proposed Mamba-Style Transformer (MST) featuring Mamba-Style Attention (MSA). 
    }
    \label{fig: Block_Comparison}
\end{figure}

The Transformer \cite{zamir2022restormer}, illustrated in Fig. \ref{fig: Block_Comparison}(a), consists of two primary components: Multi-Dconv Head Transposed Attention (MDTA) and a Gated-Dconv Feed-Forward Network (GDFN). 
In contrast, Mamba, shown in Fig. \ref{fig: Block_Comparison}(b), builds on the State Space Model (SSM) \cite{mamba}. 
However, they all have certain limitations in image restoration. 
Inspired by MILA \cite{han2024demystify} that analyzed Mamba against linear attention Transformers, we incorporate selective state modulation, a lightweight gating mechanism, into the Transformer to better preserve fine-grained local details while maintaining efficient global context modeling. 
This enhanced architecture is named the Mamba-Style Transformer (MST). 

Depicted in Fig. \ref{fig: Block_Comparison}(c), each MST first processes features with the Mamba-Style Attention (MSA), followed by refinement through GDFN, where both stages use residual connections, and generate final output $\hat{X}$. MSA is the key component, integrating a Mamba-inspired gating mechanism into MDTA for selective state modulation. Specifically, given a normalized input feature map $X \in \mathbb{R}^{H\times W\times C}$, MSA first applies a linear projection $W_l^1(\cdot)$. 
The resulting features pass through a depth-wise separable convolution, denoted as $W_d$, which aggregates spatial context while preserving channel dimensions. 
A sigmoid activation $\sigma(\cdot)$ then produces a soft mask in $[0,1]$ that modulates the subsequent attention operation $MDTA(\cdot)$ from Restormer \cite{zamir2022restormer}. 
Formally, the output $X_1$ is given by:
\begin{equation}
X_1 = (MDTA(\sigma(W_dW_l^1X)))
\label{eq:x1}
\end{equation}

In parallel, a secondary linear projection $W_l^2(\cdot)$ computes complementary features. 
These two features are combined via element‐wise multiplication and then passed through a final projection $W_l^3(\cdot)$ to refine channel‐level responses further and ensure that only salient features pass to the feed-forward network, yielding the MSA output as: 
\begin{equation}
MSA(X)=W_l^3(X_1\odot \sigma(W_l^2X))
\label{eq:msa}
\end{equation}

The sigmoid gate $\sigma(W_l^2X)$ in MSA suppresses activations corresponding to degraded tokens, achieving selective state modulation. 
It allows the attention branch to proactively suppress the degradation information and focus more effectively on regions rich in local fine-grained details. 

\subsection{Adaptive Degradation Expert Collaboration}

Leveraging the natural alignment of Mixture-of-Experts (MoE) with all-in-one restoration tasks through dynamic expert selection, we propose the Adaptive Degradation Expert Collaboration (ADEC). ADEC consists of two key components: Degradation-Aware Contextual Priors and Degradation-Guided Expert Aggregation. 

\begin{figure}[t]
    \centering
    \includegraphics[width=0.95\linewidth]{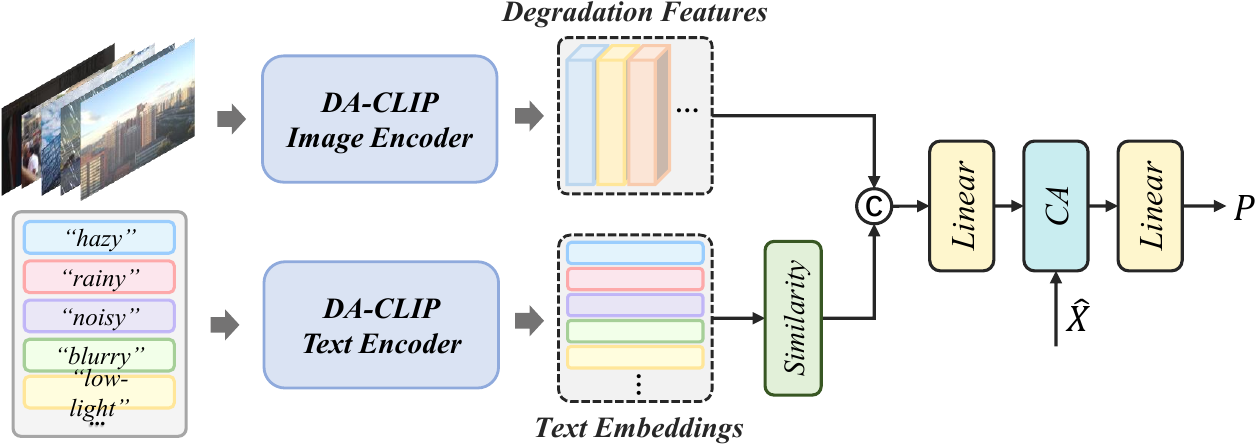}
    \caption{
    The architecture of the Degradation-Aware Contextual Priors (DACP). DACP leverages pre-trained DA-CLIP to extract degradation-aware priors, which guide the generation of more accurate routing weights.
    }
    \label{fig: DACP}
\end{figure}

\textbf{Degradation-Aware Contextual Priors.} As shown in Fig. \ref{fig: DACP}, the Degradation-Aware Contextual Priors (DACP) module first employs the pre-trained DA-CLIP image encoder and text encoder to extract the degradation features $D(I)$ from the input image $I$ and embeddings from a predefined set of textual descriptors (e.g., ``rainy'', ``hazy'', ``noisy''). The degradation similarity $S(I, text)$ is then computed based on these features. 
Degradation features and similarity are concatenated along the channel axis to form the degradation priors $P' = [D(I); S(I, text)]$. 
Since the degradation features lack spatial information, we introduce the cross-attention mechanism to achieve spatially adaptive alignment and deep fusion between the degradation priors $P'$ and the input feature map $\hat{X}$, thereby generating the position-aware degradation feature map $P$ as: 
\begin{equation}
P = W_p^2 \cdot CA(W_p^1P', \hat{X})
\label{eq:p}
\end{equation}
where $W_p^1$ and $W_p^2$ are both linear layers. 
This integration provides the subsequent routing module with precise and spatially corresponding degradation information.

\begin{figure}[t]
    \centering
    \includegraphics[width=0.95\linewidth]{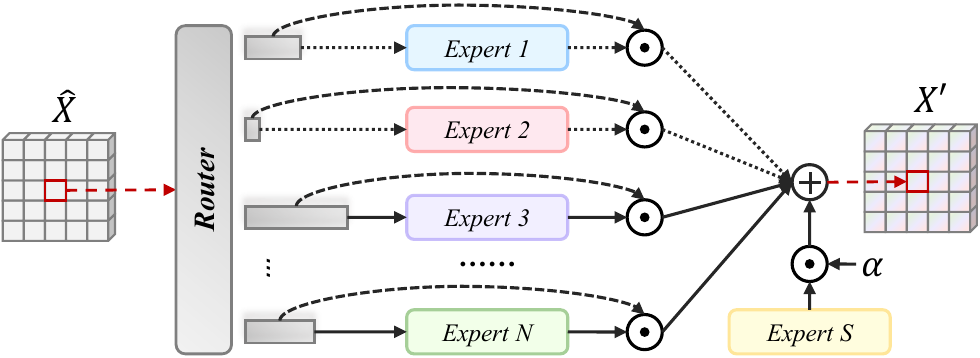}
    \caption{
    Architecture of the Degradation-Guided Expert Aggregation (DGEA). 
    For each pixel in the feature map, DGEA selects the $K$ most relevant experts based on routing weights and performs degradation-specific processing in collaboration with the shared expert. 
    }
    \label{fig: DGEA}
\end{figure}

\textbf{Degradation-Guided Expert Aggregation.} As shown in Fig. \ref{fig: DGEA}, the Degradation-Guided Expert Aggregation (DGEA) module selects specialized experts and the shared expert from an expert library $E=\{E_{1}, E_{2},..., E_{N}, E_{{S}}\}$ to restoration tasks. Here, $N$ denotes the number of specialized experts, and $E_{S}$ represents the shared expert. Routing scores are computed for each spatial position by first concatenating the position-aware degradation feature map $P$ with the layer-normalized feature $LN(\hat{X})$, then projecting the result using a learnable weight $W_r$, and finally applying a softmax to obtain normalized routing probabilities: 
\begin{equation}
score' = softmax(W_r[P;LN(\hat{X})])
\label{eq:score'}
\end{equation}
where $score'_i \in [0,1]$ and only the specialized Experts part is considered. 
We designate the expert with the highest index in $E$ as the shared expert and require its guaranteed selection, so we append a fixed high-confidence score of $1$ to the last column of $score'$. 
The scores and indices for the selected top-$K$ specialized experts and the shared expert are thus: 
\begin{equation}
score, id = top_{K+1}(softmax([score';1]))
\label{eq:score}
\end{equation}
Here, $top_{K+1}(\cdot)$ retains only the $K$ highest‐scoring specialized experts and the shared expert. 
For each pixel $(i,j)$, the outputs of the most relevant experts $\{E_{id_1}, E_{id_2},..., E_{id_K}, E_{id_{K+1}}\}$ are weighted by their corresponding routing scores and summed to produce the mixed‐expert feature: 
\begin{equation}
X_{i,j}'=\sum_{k=1}^{K+1} score_k(\hat{X})\cdot E_{id_k}(X_{i,j})
\label{eq:x'}
\end{equation}
Subsequently, $X'$ is passed through a depthwise-separable convolution $W_d'$ and a MST block $MST(\cdot)$ for feature fusion, and the fused features undergo cross-attention $CA(\cdot)$ with the original input  $X$ to produce the final output $\widetilde{\hat{X}}$ of ADEC, which can be written as: 
\begin{equation}
\widetilde{X}=CA(\hat{X},MST(W_d'X'))
\label{eq:titlex}
\end{equation}
This architecture strategically leverages the shared expert to preserve fundamental structural information and degradation-agnostic features, while enabling specialized experts to dynamically focus on distinct degradation-specific variations. 

\subsection{Loss Functions}

During training, our loss function consists of multiple components, with the primary one being the Charbonnier loss, a differentiable variant of the L1 loss. This formulation mitigates potential gradient instability caused by the non-differentiability of the standard L1 loss at zero. It is defined as follows: 
\begin{equation}
\ell_{charb} = \frac{1}{N}\sum_{i=1}^{N}\sqrt{\Vert y_i-\hat{y_i}\Vert+\epsilon^2}
\label{eq:lcharb}
\end{equation}
where $y_i$ represents the target ground-truth pixel value, $\hat{y}_i$ denotes the predicted pixel value from the model, $N$ indicates the total number of pixels involved in the calculation, $\Vert \cdot \Vert$ signifies the squared Euclidean norm, and $\epsilon$ is a predefined small positive constant. 

To alleviate the imbalance issue caused by expert selection bias, where certain experts become over-activated while others remain underutilized, we introduce a load-balancing loss function. 
This loss simultaneously encourages balance in two dimensions: the sum of experts' raw confidence weights and their actual selection frequencies. 
Specifically, for a confidence weight map $\mathcal{W} \in \mathbb{R}^{N \times H \times W}$, we first compute the total raw confidence weight $W_{n}=\sum_{i=1}^{W}\sum_{j=1}^{H}\mathcal{W}(n,i,j)$ for each expert $n=1,2,...,N$. 
This value represents the aggregate confidence assigned to expert $n$ across all spatial positions based on the model's raw weights. 
Subsequently, hard routing decisions are generated from the raw weights, yielding a binary selection tensor $\hat{\mathcal{W}}$ (where elements are 0 or 1). 
The total selected number of times each expert $n$ is $S_{n}=\sum_{i=1}^{W}\sum_{j=1}^{H}\hat{\mathcal{W}}(n,i,j)$. 
This value directly reflects the expert's actual workload. 
The load balancing loss is then defined as the sum of the squared coefficients of variation, measuring the dispersion of these two statistical sequences: 
\begin{equation}
\ell_{balance} = \frac{\sigma_{W}}{\mu_{W}^{2} + \epsilon} + \frac{\sigma_{S}}{\mu_{S}^{2} + \epsilon}
\label{eq:lbalance}
\end{equation}
where $\sigma_{W}$ and $\mu_{W}$ are the standard deviation and mean, respectively, of the sequence of total raw weights $(W_1, W_2,..., W_N)$ across all experts; $\sigma_{S}$ and $\mu_{S}$ are the standard deviation and mean, respectively, of the sequence of actual selection counts $(S_1, S_2,..., S_N)$. 
By jointly minimizing these two squared coefficient of variation terms, this loss effectively drives the model to learn parameters that promote a fairer global allocation of both the raw confidence and the actual workload among experts via the routing mechanism.

Furthermore, to enhance structural consistency in the frequency domain, we incorporate an FFT loss that measures the discrepancy between ground-truth and predicted images in Fourier space. 
This loss operates on the concatenated real and imaginary components of the Fourier Transform, defined as: 
\begin{equation}
\ell_{FFT} = \frac{1}{N}\sum_{i=1}^{N} \vert F(y_i) - F(\hat{y_i}) \vert
\label{eq:fft}
\end{equation}
where $F(x) = [Re(FFT(x));Re(FFT(x))]$ transforms an image into its Fourier domain representation. 

The overall training objective combines these components through a weighted summation: 
\begin{equation}
\ell_{total} = \ell_{charb} + \lambda_1\ell_{balance} + \lambda_2\ell_{FFT}
\label{eq:ltotal}
\end{equation}
where $\lambda_1$ and $\lambda_2$ modulate the contribution of the load-balancing and frequency-domain losses relative to the primary Charbonnier loss. This composite loss jointly optimizes pixel-level accuracy, fairness in expert allocation, and spectral fidelity throughout the optimization process. 

%% file: sec/4_experiments.tex
\section{Experiments}

To validate the effectiveness of M\textsuperscript{2}IR, we conduct comprehensive experiments across diverse all-in-one settings, encompassing eight synthetic datasets and one real-world dataset. 

\subsection{Experiment Setting}

\begin{table*}[t]
  \centering
  \caption{Quantitative Comparison of recent general and all-in-one methods on three degradations. The \textbf{best} values are highlighted in bold. }
  \begin{footnotesize}{
    \begin{tabular}{c|l|l|c|c|c|c|c|c}
    \toprule    
    \multirow{2}{*}{} & \multirow{2}{*}{\textbf{Method}} & \multirow{2}{*}{\textbf{Venue}} & \textbf{Dehazing} & \textbf{Deraining} & \multicolumn{3}{c|}{\textbf{Denoising on BSD68} \cite{martin2001database}} & \multirow{2}{*}{\textbf{Average}} \\
      &  &  & \multicolumn{1}{p{5.8em}|}{\textbf{on SOTS} \cite{li2018benchmarking}} & \multicolumn{1}{p{7.8em}|}{\textbf{on Rain100L} \cite{yang2017deep}} & $\sigma=15$ & $\sigma=25$ & $\sigma=50$ & \\
    \midrule
    \multirow{4}{*}{\rotatebox[origin=c]{90}{\textit{\textbf{General}}}} 
    & Restormer \cite{zamir2022restormer} & CVPR'22 & 27.78\;/\;0.958 & 33.78\;/\;0.958 & 33.72\;/\;0.930 & 30.67\;/\;0.865 & 27.63\;/\;0.792 & 30.75\;/\;0.901  \\ 
    & NAFNet \cite{chen2022simple} & ECCV'22 & 24.11\;/\;0.928 & 33.64\;/\;0.956 & 33.03\;/\;0.918 & 30.47\;/\;0.865 & 27.12\;/\;0.754 & 29.67\;/\;0.844  \\ 
    & FSNet \cite{cui2023image} & TPAMI'23 & 29.14\;/\;0.968 & 35.61\;/\;0.969 & 33.81\;/\;0.930 & 30.84\;/\;0.872 & 27.69\;/\;0.792 & 31.42\;/\;0.905  \\ 
    & MambaIR \cite{guo2024mambair} & ECCV'24 & 29.57\;/\;0.970 & 35.42\;/\;0.969 & 33.88\;/\;0.931 & 30.95\;/\;0.874 & 27.74\;/\;0.793 & 31.51\;/\;0.907 \\
    \midrule
    \multirow{13}{*}{\rotatebox[origin=c]{90}{\textit{\textbf{All-in-One}}}} 
    & AirNet \cite{li2022all} & CVPR'22 & 27.94\;/\;0.962 & 34.90\;/\;0.967 & 33.92\;/\;0.933 & 31.26\;/\;0.888 & 28.00\;/\;0.797 & 31.20\;/\;0.910 \\
    & IDR \cite{zhang2023ingredient} & CVPR'23 & 29.87\;/\;0.970 & 36.03\;/\;0.971 & 33.89\;/\;0.931 & 31.32\;/\;0.884 & 28.04\;/\;0.798 & 31.83\;/\;0.911 \\
    & PromptIR \cite{potlapalli2023promptir} & NeurIPS'23 & 30.58\;/\;0.974 & 36.37\;/\;0.972 & 33.98\;/\;0.933 & 31.31\;/\;0.888 & 28.06\;/\;0.799 & 32.06\;/\;0.913 \\
    & Gridformer \cite{wang2024gridformer} & IJCV'24 & 30.37\;/\;0.970 & 37.15\;/\;0.972 & 33.93\;/\;0.931 & 31.37\;/\;0.887 & 28.11\;/\;0.801 & 32.19\;/\;0.912 \\
    & InstructIR-3D \cite{conde2024high} & ECCV'24 & 30.22\;/\;0.959 & 37.98\;/\;0.978 & 34.15\;/\;0.933 & 31.52\;/\;0.890 & 28.30\;/\;0.804 & 32.43\;/\;0.913 \\
    & Perceive-IR \cite{zhang2025perceive} & TIP'25 & 30.87\;/\;0.975 & 38.29\;/\;0.980 & 34.13\;/\;0.934 & 31.53\;/\;0.890 & 28.31\;/\;0.804 & 32.63\;/\;0.917 \\
    & Pool-AIO \cite{cui2025exploring} & TIP'25 & 30.94\;/\;0.980 & 38.54\;/\;0.983 & 34.10\;/\;0.935 & 31.45\;/\;0.892 & 28.18\;/\;0.803 & 32.64\;/\;0.919 \\
    & AdaIR \cite{cui2025adair} & ICLR'25 & 31.06\;/\;0.980 & 38.64\;/\;0.983 & 34.12\;/\;0.935 & 31.45\;/\;0.892 & 28.19\;/\;0.802 & 32.69\;/\;0.918 \\
    & MoCE-IR \cite{zamfir2025complexity} & CVPR'25 & 31.34\;/\;0.979 & 38.57\;/\;0.984 & 34.11\;/\;0.932 & 31.45\;/\;0.888 & 28.18\;/\;0.800 & 32.73\;/\;0.917 \\
    &MEASNet \cite{yu2024multi} & TCSVT'25 & 31.61\;/\;0.981 & 39.00\;/\;0.985 & 34.12\;/\;0.935 & 31.46\;/\;0.892 & 28.19\;/\;0.803 & 32.85\;/\;0.919 \\
    &MIRAGE \cite{ren2026manifold} & ICLR'26 & 31.86\;/\;0.981 & 38.94\;/\;0.985 & 34.12\;/\;0.935 & 31.46\;/\;0.891 & 28.19\;/\;0.803 & 32.91\;/\;0.919 \\
    &BaryIR \cite{tang2026learning} & TPAMI'26 & 31.40\;/\;0.980 & 39.02\;/\;\textbf{0.985} & 34.16\;/\;0.935 & 31.54\;/\;0.892 & 28.25\;/\;0.802 & 32.86\;/\;0.919 \\
    \cmidrule{2-9}
    & M\textsuperscript{2}IR & Ours & \textbf{32.63}\;/\;\textbf{0.984} & \textbf{39.36}\;/\;\textbf{0.986} & \textbf{34.23}\;/\;\textbf{0.937} & \textbf{31.59}\;/\;\textbf{0.895} & \textbf{28.33}\;/\;\textbf{0.808} & \textbf{33.23}\;/\;\textbf{0.922} \\
    \bottomrule
    \end{tabular}
    }
    \end{footnotesize}
  \label{tab:3dir}
\end{table*}

\begin{figure*}[t]
    \centering
    \includegraphics[width=1\linewidth]{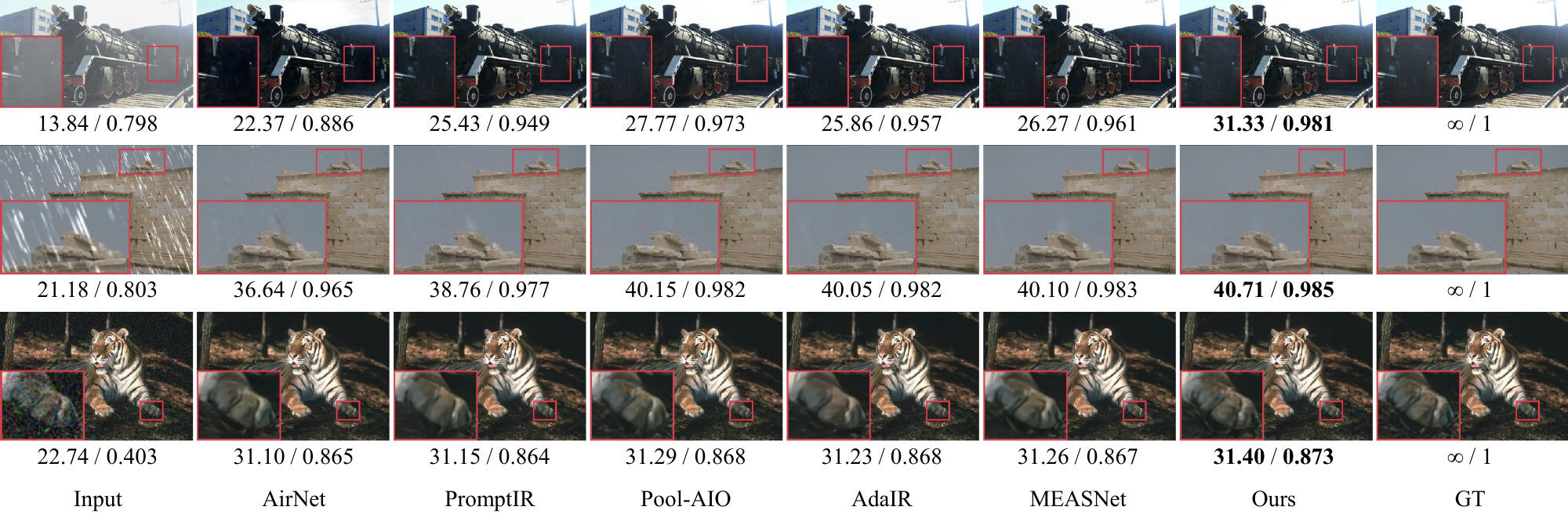}
    \caption{Visual comparisons of M\textsuperscript{2}IR with state-of-the-art all-in-one methods on three degradations, with magnified views of red-boxed regions provided for better visualization. The task types from top to bottom are dehazing, deraining, and denoising, respectively. }
    \label{fig:3d_visual}
\end{figure*}

\begin{table*}[t]
  \centering
  \caption{Quantitative Comparison of recent general and all-in-one methods on five degradations. Denoising results are reported for the noise level $\sigma=25$.}
  \begin{footnotesize}{
  \resizebox{\textwidth}{!}{
    \begin{tabular}{c|l|l|c|c|c|c|c|c}
    \toprule
    \multirow{2}{*}{} & \multirow{2}{*}{\textbf{Method}} & \multirow{2}{*}{\textbf{Venue}} & \textbf{Dehazing} & \textbf{Deraining} & \textbf{Denoising} & \textbf{Deblurring} & \textbf{Low-Light} & \multirow{2}{*}{\textbf{Average}} \\
      &   &   & \multicolumn{1}{p{5.7em}|}{\textbf{on SOTS} \cite{li2018benchmarking}} & \multicolumn{1}{p{7.45em}|}{\textbf{on Rain100L} \cite{yang2017deep}} & \multicolumn{1}{p{6.15em}|}{\textbf{on BSD68} \cite{martin2001database}} & \multicolumn{1}{p{6em}|}{\textbf{on GoPro} \cite{nah2017deep}} & \multicolumn{1}{p{5.3em}|}{\textbf{on LOL} \cite{wei2018deep}} &  \\
    \midrule
    \multirow{4}{*}{\rotatebox[origin=c]{90}{\textit{\textbf{General}}}} 
    & Restormer \cite{zamir2022restormer} & CVPR'22 & 24.09\;/\;0.927 & 34.81\;/\;0.962 & 31.49\;/\;0.884 & 27.22\;/\;0.829 & 20.41\;/\;0.806 & 27.60\;/\;0.881  \\ 
    & NAFNet \cite{chen2022simple} & ECCV'22 & 25.23\;/\;0.939 & 35.56\;/\;0.967 & 31.02\;/\;0.883 & 26.53\;/\;0.808 & 20.49\;/\;0.809 & 27.76\;/\;0.881  \\ 
    & FSNet \cite{cui2023image} & TPAMI'23 & 25.53\;/\;0.943 & 36.07\;/\;0.968 & 31.33\;/\;0.883 & 28.32\;/\;0.869 & 22.29\;/\;0.829 & 28.71\;/\;0.898  \\ 
    & MambaIR \cite{guo2024mambair} & ECCV'24 & 25.81\;/\;0.944 & 36.55\;/\;0.971 & 31.41\;/\;0.884 & 28.61\;/\;0.875 & 22.49\;/\;0.832 & 28.97\;/\;0.901 \\
    \midrule
    \multirow{13}{*}{\rotatebox[origin=c]{90}{\textit{\textbf{All-in-One}}}} 
    & AirNet \cite{li2022all} & CVPR'22 & 21.04\;/\;0.884 & 32.98\;/\;0.951 & 30.91\;/\;0.882 & 24.35\;/\;0.781 & 18.18\;/\;0.735 & 25.49\;/\;0.846 \\
    & IDR \cite{zhang2023ingredient} & CVPR'23 & 25.24\;/\;0.943 & 35.63\;/\;0.965 & \textbf{31.60}\;/\;0.887 & 27.87\;/\;0.846 & 21.34\;/\;0.826 & 28.34\;/\;0.893 \\
    & PromptIR \cite{potlapalli2023promptir} & NeurIPS'23 & 26.54\;/\;0.949 & 36.37\;/\;0.970 & 31.47\;/\;0.886 & 28.71\;/\;0.881 & 22.68\;/\;0.832 & 29.15\;/\;0.904 \\
    & Gridformer \cite{wang2024gridformer} & IJCV'24 & 26.79\;/\;0.951 & 36.61\;/\;0.971 & 31.45\;/\;0.885 & 29.22\;/\;0.884 & 22.59\;/\;0.831 & 29.33\;/\;0.904 \\
    & InstructIR-5D \cite{conde2024high} & ECCV'24 & 27.10\;/\;0.956 & 36.84\;/\;0.973 & 31.40\;/\;0.887 & 29.40\;/\;0.886 & 23.00\;/\;0.836 & 29.55\;/\;0.907 \\
    & Perceive-IR \cite{zhang2025perceive} & TIP'25 & 28.19\;/\;0.964 & 37.25\;/\;0.977 & 31.44\;/\;0.887 & 29.46\;/\;0.886 & 22.88\;/\;0.833 & 29.84\;/\;0.909 \\
    & Pool-AIO \cite{cui2025exploring} & TIP'25 & 30.25\;/\;0.977 & 37.85\;/\;0.981 & 31.24\;/\;0.887 & 27.66\;/\;0.844 & 22.66\;/\;0.841 & 29.93\;/\;0.906 \\
    & AdaIR \cite{cui2025adair} & ICLR'25 & 30.53\;/\;0.978 & 38.02\;/\;0.981 & 31.35\;/\;0.889 & 28.12\;/\;0.858 & 23.00\;/\;0.845 & 30.20\;/\;0.910 \\
    & MoCE-IR \cite{zamfir2025complexity} & CVPR'25 & 30.48\;/\;0.974 & 38.04\;/\;0.982 & 31.34\;/\;0.887 & \textbf{30.05}\;/\;\textbf{0.899} & 23.00\;/\;0.852 & 30.58\;/\;0.919 \\
    &MEASNet \cite{yu2024multi} & TCSVT'25 & 31.05\;/\;0.980 & 38.32\;/\;0.982 & 31.40\;/\;0.888 & 29.41\;/\;0.890 & 23.00\;/\;0.845 & 30.64\;/\;0.917 \\
    &MIRAGE \cite{ren2026manifold} & ICLR'26 & 31.45\;/\;0.980 & 38.92\;/\;\textbf{0.985} & 31.41\;/\;0.892 & 28.10\;/\;0.858 & 23.59\;/\;0.858 & 30.68\;/\;0.914 \\
    &BaryIR \cite{tang2026learning} & TPAMI'26 & 31.20\;/\;0.979 & 38.10\;/\;0.982 & 31.43\;/\;0.891 & 29.51\;/\;0.889 & 23.37\;/\;0.854 & 30.72\;/\;0.919 \\
    \cmidrule{2-9}
    & M\textsuperscript{2}IR & Ours & \textbf{31.63}\;/\;\textbf{0.981} & \textbf{39.28}\;/\;\textbf{0.985} & 31.53\;/\;\textbf{0.893} & 29.09\;/\;0.878 & \textbf{23.80}\;/\;\textbf{0.864} & \textbf{31.06}\;/\;\textbf{0.920}  \\
    \bottomrule
    \end{tabular}
    }
    }
    \end{footnotesize}
  \label{tab:5dir}
\end{table*}

\begin{table*}[t]
  \centering
  \caption{Quantitative Comparison of different methods on CDD11 \cite{guo2024onerestore} Dataset. * denotes the model has been retrained.}
  \setlength{\tabcolsep}{0.1cm}
    \begin{footnotesize}{
    \resizebox{\textwidth}{!}{
    \begin{tabular}{l|cccc|ccccc|cc|c}
    \toprule
    \multirow{2}{*}{\textbf{Method}} & \multicolumn{4}{c|}{\textbf{Single}} & \multicolumn{5}{c|}{\textbf{Double}} & \multicolumn{2}{c|}{\textbf{Triple}} & \multirow{2}{*}{\textbf{Average}} \\ 
      & \textbf{L} & \textbf{H} & \textbf{R} & \textbf{S} & \textbf{L+H} & \textbf{L+R} & \textbf{L+S} & \textbf{H+R} & \textbf{H+S} & \textbf{L+H+R} & \textbf{L+H+S} &  \\
    \midrule
    AirNet \cite{li2022all} & 24.83/0.778 & 24.21/0.951 & 26.55/0.891 & 26.79/0.919 & 23.23/0.779 & 22.82/0.710 & 23.29/0.723 & 22.21/0.868 & 23.29/0.901 & 21.80/0.708 & 22.24/0.725 & 23.75/0.814 \\
    PromptIR \cite{potlapalli2023promptir} & 26.32/0.805 & 26.10/0.969 & 31.56/0.946 & 31.53/0.960 & 24.49/0.789 & 25.05/0.771 & 24.51/0.761 & 24.54/0.924 & 23.70/0.925 & 23.74/0.752 & 23.33/0.747 & 25.90/0.850 \\
    WeatherDiff \cite{ozdenizci2023restoring} & 23.58/0.763 & 21.99/0.904 & 24.85/0.885 & 24.80/0.888 &  21.83/0.756 & 22.69/0.730 & 22.12/0.707 & 21.25/0.868 & 21.99/0.868 & 21.23/0.716 & 21.04/0.698 & 22.49/0.799 \\
    WGWS-Net \cite{zhu2023learning} & 24.39/0.774 & 27.90/0.982 & 33.15/0.964 & 34.43/0.973 & 24.27/0.800 & 25.06/0.772 & 24.60/0.765 & 27.23/0.955 & 27.65/0.960 & 23.90/0.772 & 23.97/0.771 & 26.96/0.863 \\
    OneRestore \cite{guo2024onerestore} & 26.48/0.826 & 32.52/0.990 & 33.40/0.964 & 34.31/0.973 & 25.79/0.822 & 25.58/0.799 & 25.19/0.789 & 29.99/0.957 & 30.21/0.964 & 24.78/0.788 & 24.90/0.791 & 28.47/0.878 \\
    AdaIR* \cite{cui2025adair} & 26.88/0.821 & 31.60/0.987 & 33.84/0.962 & 34.65/0.974 & 25.69/0.811 & 25.90/0.793 & 25.69/0.783 & 29.38/0.955 & 28.95/0.961 & 24.82/0.778 & 25.04/0.778 & 28.40/0.873 \\
    MoCE-IR \cite{zamfir2025complexity} & 27.26/0.824 & 32.66/0.990 & 34.31/0.970 & 35.91/0.980 & 26.24/0.817 & 26.25/0.800 & 26.04/0.793 & 29.93/0.964 & 30.19/0.970 & 25.41/0.789 & 25.39/0.790 & 29.05/0.881 \\
    MIRAGE \cite{ren2026manifold} & 27.41/0.833 & 33.12/\textbf{0.992} & 34.66/0.971 & 35.98/\textbf{0.981} & 26.55/0.828 & 26.53/0.810 & 26.33/0.803 & 30.32/0.965 & 30.27/0.969 & 25.59/0.801 & 25.86/0.799 & 29.33/0.887 \\
    \midrule
    M\textsuperscript{2}IR & \textbf{27.67}/\textbf{0.836} & \textbf{34.84}/\textbf{0.992} & \textbf{35.39}/\textbf{0.974} & \textbf{37.11}/\textbf{0.981} & \textbf{26.87}/\textbf{0.831} & \textbf{26.77}/\textbf{0.814} & \textbf{26.62}/\textbf{0.808} & \textbf{31.89}/\textbf{0.970} & \textbf{31.78}/\textbf{0.973} & \textbf{26.07}/\textbf{0.805} & \textbf{26.15}/\textbf{0.804} & \textbf{30.11}/\textbf{0.890} \\
    \bottomrule
    \end{tabular}
    }}
    \end{footnotesize}    
  \label{tab:cdd11}
\end{table*}

\begin{figure*}[!htp]
    \centering
    \includegraphics[width=1\linewidth]{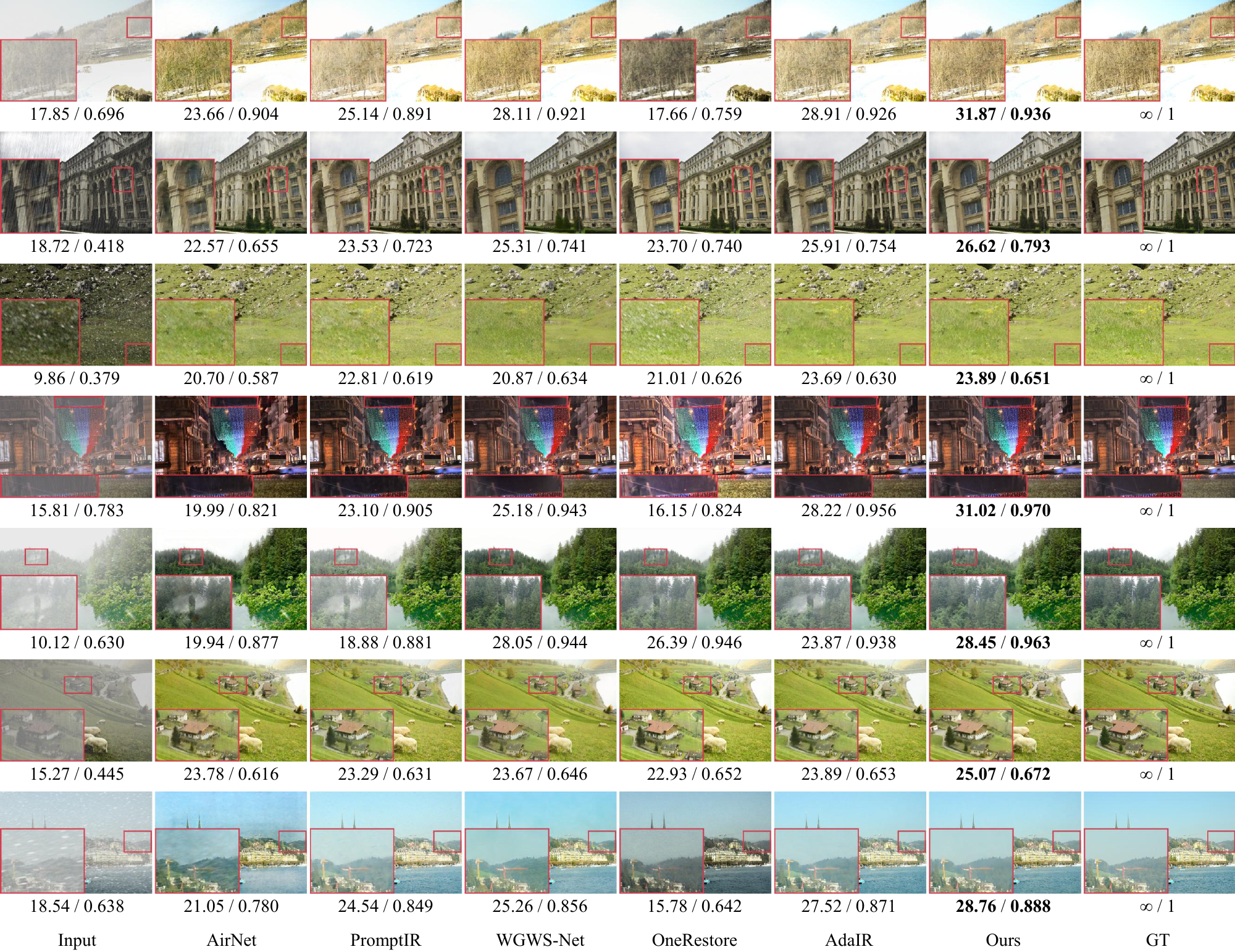}
    \caption{Visual comparisons of M\textsuperscript{2}IR with state-of-the-art all-in-one methods on CDD11 \cite{guo2024onerestore} dataset, with magnified views of red-boxed regions provided for better visualization. The composite degradation types from top to bottom are L+H, L+R, L+S, H+R, H+S, L+H+R, and L+H+S, respectively. 
    }
    \label{fig:cdd11_visual}
\end{figure*}

\textbf{Datasets.} We closely follow existing works \cite{yu2024multi, cui2025adair}. 
For dehazing, we select the SOTS from the RESIDE (outdoor) \cite{li2018benchmarking}  dataset.
For draining, we adopt the Rain100L \cite{yang2017deep} dataset. 
For denoising, we synthetically generate noisy images by adding Gaussian noise with noise levels $\sigma \in \{15,25,50\}$ to the clean images from the WED \cite{ma2016waterloo} and BSD400 \cite{martin2001database} datasets. 
Evaluation is performed on the BSD68 \cite{martin2001database} dataset.
For deblurring, we employ the GoPro \cite{nah2017deep} dataset. 
For low-light enhancement, we use the LOL \cite{wei2018deep} dataset. 
To construct our all-in-one restoration model for all tasks, we merge these datasets under multi-degradation settings, including ``Haze, Rain, Noise'' and ``Haze, Rain, Noise, Blur, Low-light''. 
For the composite degradation, we use the CDD11 \cite{guo2024onerestore} dataset, which contains 11 types of degradation scenarios. 
Each type is constructed from four basic degradations: low-light (L), hazy (H), rainy (R), and snowy (S), which can be used individually or in various combinations. 
The degradation types include L, H, R, S, L+H, L+R, L+S, H+R, H+S, L+H+R, and L+H+S, covering a comprehensive range of single and composite degradation conditions for image restoration tasks. 
For real-world degradation, we use the WeatherBench \cite{guan2025weatherbench} dataset, which serves as the first comprehensive all-in-one benchmark for real-world adverse weather image restoration. It provides accurately aligned degraded-clean image pairs captured under diverse natural conditions. This dataset encompasses various weather conditions such as haze, rain, and snow, which are artificially generated in the real world, covering both daytime and nighttime scenes. 

\textbf{Implementation Details.} 
Our M\textsuperscript{2}IR is implemented using the PyTorch framework and trained on two NVIDIA GeForce RTX 4090 GPUs. 
During the training phase, we randomly crop images into $128 \times 128$ patches and apply data augmentation through random flipping and rotation. 
The network is optimized using an AdamW optimizer ($\beta_1=0.9$, $\beta_2=0.999$) and a linear warmup cosine annealing scheduler with an initial learning rate of $2\times10^{-4}$. 

\subsection{Comparison with State-of-the-art Methods}

For objective performance evaluation, unless otherwise specified, we mainly adopt widely recognized image quality assessment metrics: Peak Signal-to-Noise Ratio (PSNR) and Structural Similarity Index Measure (SSIM). 
Both metrics follow a higher-is-better evaluation principle. 

\textbf{Comparison on Three Degradations.} We conduct comprehensive comparisons with several state-of-the-art all-in-one methods AirNet \cite{li2022all}, IDR \cite{zhang2023ingredient}, PromptIR \cite{potlapalli2023promptir}, Gridformer \cite{wang2024gridformer}, InstructIR \cite{conde2024high}, Perceive-IR \cite{zhang2025perceive}, Pool-AIO \cite{cui2025exploring}, AdaIR \cite{cui2025adair}, MoCE-IR \cite{zamfir2025complexity}, MEASNet \cite{yu2024multi}, MIRAGE \cite{ren2026manifold}, and BaryIR \cite{tang2026learning}, alongside general image restoration methods Restormer \cite{zamir2022restormer}, NAFNet \cite{chen2022simple}, FSNet \cite{cui2023image}, and MambaIR \cite{guo2024mambair}. 
These comparisons span three degradation tasks, including dehazing, deraining, and multi-level denoising, where the noise levels are set to $\sigma=15$, $\sigma=25$, and $\sigma=50$. 
As quantitatively demonstrated in Tab.~\ref{tab:3dir}, our M\textsuperscript{2}IR achieves the highest PSNR and SSIM scores on every individual task, representing an improvement of +0.32 dB PSNR and +0.003 SSIM over the previous best method MIRAGE on average. 
In addition to the quantitative results, we further provide qualitative comparisons with competing approaches. 
As illustrated in Fig.~\ref{fig:3d_visual}, we present visual comparisons of several representative methods, and our method can preserve more faithful fine-grained local details, yielding reconstructions that are visually closer to the ground truth. 
This demonstrates that M\textsuperscript{2}IR not only achieves superior numerical performance but also delivers perceptually consistent restorations across diverse degradations. 

\textbf{Comparison on Five Degradations.} To further verify the performance of our method, we follow the common practices in existing research. 
We applied it to five image restoration tasks, including dehazing, deraining, denoising, deblurring, and low-light image enhancement. 
As detailed in Tab.~\ref{tab:5dir}, our approach establishes new state-of-the-art results in dehazing, deraining, and low-light enhancement, while achieving competitive leading performance in denoising and deblurring. 
Although slightly trailing in deblurring, our method attains the highest overall average PSNR and SSIM scores, significantly surpassing BaryIR by 0.34 dB in PSNR, demonstrating its superior effectiveness across diverse degradation scenarios. 

\textbf{Comparison on CDD11 Dataset.} To even further verify the effectiveness of the proposed method in handling composite degradations, we conduct experiments on the CDD11 \cite{guo2024onerestore} dataset, which contains both single and composite degradation types. 
We compare our approach with representative all-in-one methods, also named one-to-many, including AirNet \cite{li2022all}, PromptIR \cite{potlapalli2023promptir}, WeatherDiff \cite{ozdenizci2023restoring}, WGWS-Net \cite{zhu2023learning}, AdaIR \cite{cui2025adair}, MoCE-IR \cite{zamfir2025complexity}, and MIRAGE \cite{ren2026manifold}, as well as the one-to-composite method OneRestore \cite{guo2024onerestore}. 
As shown in Tab.~\ref{tab:cdd11}, our method achieves consistent advantages across all degradation types. 
Compared with the previous state-of-the-art MIRAGE, our method improves the average PSNR from 29.33 dB to 30.11 dB and the average SSIM from 0.887 to 0.890, indicating its superior capacity in restoring images under complex composite degradation conditions. 
To provide a more intuitive understanding of the restoration quality, we further present visual comparisons on 7 composite degradation types from the CDD11 \cite{guo2024onerestore} dataset in Fig.~\ref{fig:cdd11_visual}, where our method produces reconstructions with more fine-grained local details, proactively suppressing the interference of multiple degradations and yielding results that are visually more faithful to the reference images.

\begin{table*}[t]
  \centering
  \caption{Quantitative Comparison of different methods on WeatherBench \cite{guan2025weatherbench} Dataset.}
  \begin{footnotesize}{
  \resizebox{\textwidth}{!}{
    \begin{tabular}{l|l|c@{\hspace{0.2cm}}c@{\hspace{0.2cm}}c@{\hspace{0.2cm}}c|c@{\hspace{0.2cm}}c@{\hspace{0.2cm}}c@{\hspace{0.2cm}}c|c@{\hspace{0.2cm}}c@{\hspace{0.2cm}}c@{\hspace{0.2cm}}c|c@{\hspace{0.2cm}}c@{\hspace{0.2cm}}c@{\hspace{0.2cm}}c}
    \toprule
    \multirow{2}{*}{\textbf{Method}} & \multirow{2}{*}{\textbf{Venue}} & \multicolumn{4}{c|}{\textbf{Dehazing}} & \multicolumn{4}{c|}{\textbf{Deraining}} & \multicolumn{4}{c|}{\textbf{Desnowing}} & \multicolumn{4}{c}{\textbf{Average}} \\
      &   & \textbf{PSNR$\uparrow$} & \textbf{SSIM$\uparrow$} & \textbf{LPIPS$\downarrow$} & \textbf{FID$\downarrow$} & \textbf{PSNR$\uparrow$} & \textbf{SSIM$\uparrow$} & \textbf{LPIPS$\downarrow$} & \textbf{FID$\downarrow$} & \textbf{PSNR$\uparrow$} & \textbf{SSIM$\uparrow$} & \textbf{LPIPS$\downarrow$} & \textbf{FID$\downarrow$} & \textbf{PSNR$\uparrow$} & \textbf{SSIM$\uparrow$} & \textbf{LPIPS$\downarrow$} & \textbf{FID$\downarrow$} \\
    \midrule
    AirNet \cite{li2022all} & CVPR'22 & 19.27  & 0.645  & 0.3829  & 134.09  & 31.56  & 0.912  & 0.2236  & 125.54  & 20.58  & 0.737  & 0.2912  & 138.57  & 23.80  & 0.764  & 0.2992  & 132.73  \\
    TransWeather \cite{valanarasu2022transweather} & CVPR'22 & 18.13  & 0.621  & 0.3970  & 123.21  & 28.59  & 0.880  & 0.2638  & 149.66  & 24.06  & 0.754  & 0.2250  & 102.99  & 23.59  & 0.752  & 0.2953  & 125.29  \\
    PromptIR \cite{potlapalli2023promptir} & NeurIPS'23 & 19.50  & 0.658  & 0.3751  & 113.55  & 32.51  & 0.915  & 0.1980  & 111.69  & 26.35  & 0.804  & 0.1951  & 84.12  & 26.12  & 0.792  & 0.2561  & 103.12  \\
    WGWS-Net \cite{zhu2023learning} & CVPR'23 & 11.78  & 0.532  & 0.5351  & 152.76  & 34.77  & 0.939  & \textbf{0.1168}  & \textbf{60.99}  & 19.39  & 0.721  & 0.2481  & 128.56  & 21.98  & 0.731  & 0.3000  & 114.10  \\
    Histoformer \cite{sun2024restoring} & ECCV'24 & 15.82  & 0.597  & 0.4371  & 128.34  & 28.87  & 0.876  & 0.2785  & 152.42  & 23.88  & 0.769  & 0.2252  & 105.82  & 22.86  & 0.747  & 0.3136  & 128.86  \\
    AdaIR \cite{cui2025adair} & ICLR'25 & 21.39  & 0.680  & 0.3506  & 110.07  & 32.81  & 0.918  & 0.1916  & 109.41  & 26.87  & 0.806  & 0.1790  & 73.48  & 27.02  & 0.801  & 0.2404  & 97.65  \\
    DiffUIR \cite{zheng2024selective} & CVPR'24 & 20.96  & 0.695  & 0.3550  & 127.54  & 33.78  & 0.931  & 0.1720  & 86.96  & 27.87  & 0.844  & 0.1619  & 68.99  & 27.54  & 0.823  & 0.2296  & 94.50  \\
    \midrule
    M\textsuperscript{2}IR & Ours & \textbf{23.63}  & \textbf{0.739}  & \textbf{0.3232}  & \textbf{109.82}  & \textbf{35.71}  & \textbf{0.943}  & 0.1597  & 78.75  & \textbf{29.52}  & \textbf{0.874}  & \textbf{0.1505}  & \textbf{62.93}  & \textbf{29.62}  & \textbf{0.852}  & \textbf{0.2111}  & \textbf{83.83}  \\
    \bottomrule
    \end{tabular}
    }}
    \end{footnotesize}
  \label{tab:weatherbench}
\end{table*}

\begin{figure*}[ht]
    \centering
    \includegraphics[width=1\linewidth]{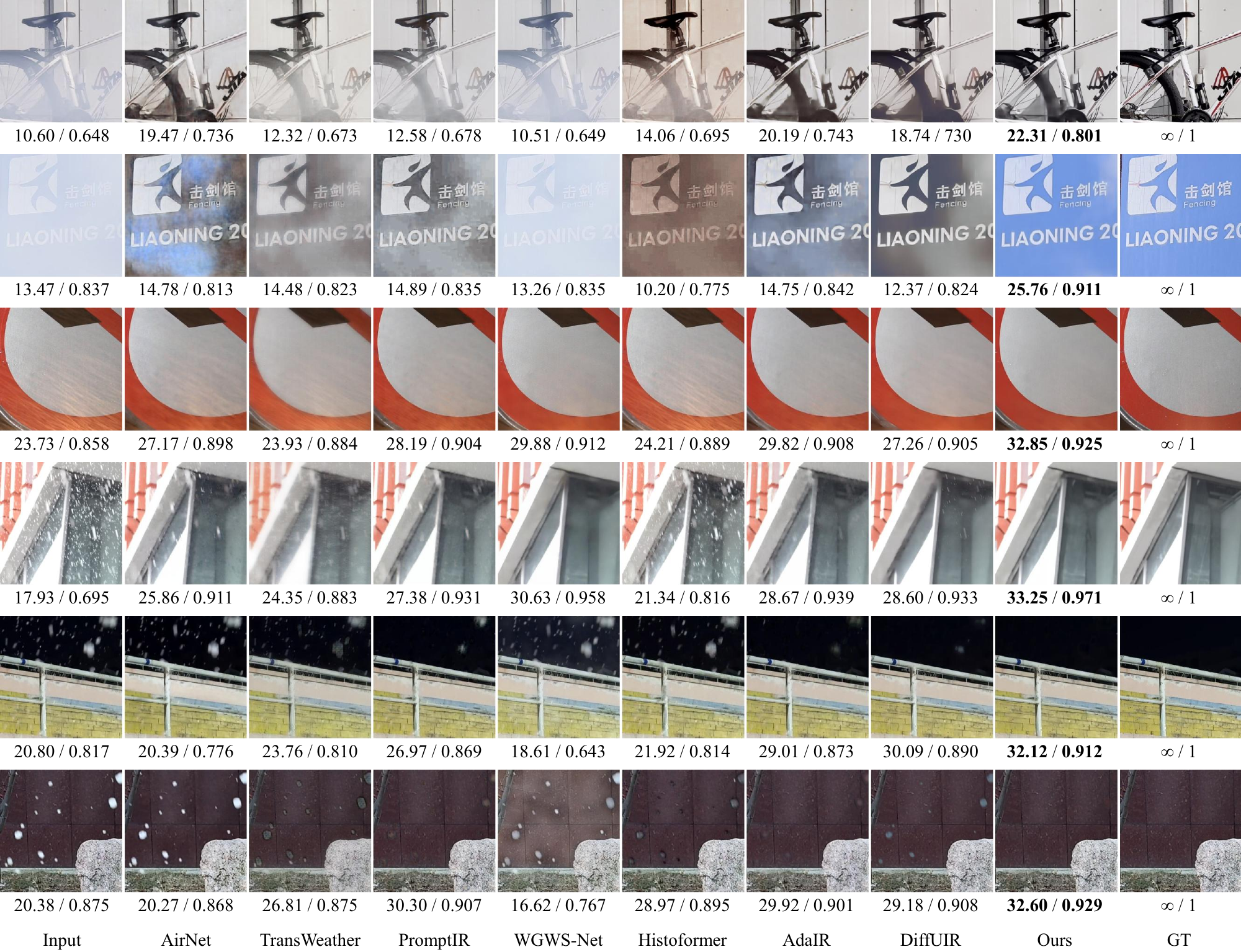}
    \caption{Visual comparisons of M\textsuperscript{2}IR with state-of-the-art all-in-one methods on WeatherBench \cite{guan2025weatherbench} dataset. The degradation types, from top to bottom, are haze (rows 1 and 2), rain (rows 3 and 4), and snow (rows 5 and 6). 
    }
    \label{fig:weatherbench_visual}
\end{figure*}

\textbf{Comparison on WeatherBench Dataset.} To comprehensively verify the performance of M\textsuperscript{2}IR on real-world datasets, we conduct experiments on the WeatherBench \cite{guan2025weatherbench} dataset. 
We directly employ the restoration results of multiple all-in-one methods (AirNet \cite{li2022all}, TransWeather \cite{valanarasu2022transweather}, PromptIR \cite{potlapalli2023promptir}, WGWS-Net \cite{zhu2023learning}, Histoformer \cite{sun2024restoring}, AdaIR \cite{cui2025adair}, and DiffUIR \cite{zheng2024selective}) from WeatherBench and recalculate the metrics using the same code.
As shown in Tab.~\ref{tab:weatherbench}, our M\textsuperscript{2}IR achieves a 2.08 dB improvement in PSNR and a 0.029 increase in SSIM compared to the previous state-of-the-art method DiffUIR. 
To further evaluate the generalization and robustness of our method, we additionally employ LPIPS and FID for evaluation, where our method also attains the best or highly competitive results, indicating its excellent perceptual quality and feature-level fidelity. 
Furthermore, the visual comparisons presented in Fig.~\ref{fig:weatherbench_visual} show that M\textsuperscript{2}IR can most effectively remove haze, rain, and snow, recovering results that are closest to the ground truth on the real-world dataset WeatherBench \cite{guan2025weatherbench}. 

\textbf{Comparison on One-by-One Restoration}

\begin{table}[t]
  \centering
  \caption{
  Quantitative Comparison of recent image restoration methods under the one-by-one settings, with each sub-table dividing methods into task-specific (upper), general-purpose (middle), and all-in-one (lower) categories. * denotes the model has been retrained.
  }
  \resizebox{\columnwidth}{!}{
    \begin{tabular}{c}
    \begin{tabular}{l|c@{\hspace{0.2cm}}c@{\hspace{0.2cm}}c|c@{\hspace{0.2cm}}c@{\hspace{0.2cm}}c|c@{\hspace{0.2cm}}c@{\hspace{0.2cm}}c}
    \toprule
    \multirow{2}{*}{\textbf{Method}} & \multicolumn{3}{c|}{\textbf{BSD68} \cite{martin2001database}} & \multicolumn{3}{c|}{\textbf{Urban100} \cite{huang2015single}} & \multicolumn{3}{c}{\textbf{Kodak24} \cite{kodak24}} \\
      & 15 & 25 & 50 & 15 & 25 & 50 & 15 & 25 & 50 \\
    \midrule
     DnCNN \cite{zhang2017beyond} & 33.90 & 31.24 & 27.95 & 32.98 & 30.81 & 27.59 & 34.60 & 32.14 & 28.95 \\
     FFDNet \cite{zhang2018ffdnet} & 33.87 & 31.21 & 27.96 & 33.83 & 31.40 & 28.05 & 34.63 & 32.13 & 28.98 \\
     ADFNet \cite{shen2023adaptive} & 34.21 & 31.60 & 28.19 & 34.50 & 32.13 & 28.71 & 34.77 & 32.22 & 29.06 \\
    \midrule
     MIRNet-v2 \cite{zamir2022learning} & 33.66 & 30.97 & 27.66 & 33.30 & 30.75 & 27.22 & 34.29 & 31.81 & 28.55 \\
     DGUNet \cite{mou2022deep} & 33.85 & 31.10 & 27.92 & 33.67 & 31.27 & 27.94 & 34.56 & 32.10 & 28.91 \\
     Restormer \cite{zamir2022restormer} & 34.03 & 31.49 & 28.11 & 33.72 & 31.26 & 28.03 & 34.78 & 32.37 & 29.08 \\
     NAFNet \cite{chen2022simple} & 33.67 & 31.02 & 27.73 & 33.14 & 30.64 & 27.20 & 34.27 & 31.80 & 28.62 \\
     FSNet \cite{cui2023image} & 34.09 & 31.55 & 28.12 & 33.88 & 31.31 & 28.07 & 34.75 & 32.38 & 29.10 \\
    \midrule
     AirNet \cite{li2022all} & 34.14 & 31.48 & 28.23 & 34.40 & 32.10 & 28.88 & 34.81 & 32.44 & 29.10 \\
     PromptIR* \cite{potlapalli2023promptir} & 34.33 & 31.70 & 28.45 & 34.83 & 32.59 & 29.50 & 35.30 & 32.86 & 29.76 \\
     Perceive-IR \cite{zhang2025perceive} & 34.38 & 31.74 & 28.53 & 34.86 & 32.55 & 29.42 & 34.84 & 32.50 & 29.16 \\
     AdaIR \cite{cui2025adair} & 34.36 & 31.72 & 28.49 & 34.96 & 32.74 & 29.70 & 35.39 & 32.94 & 29.85 \\
      M\textsuperscript{2}IR & \textbf{34.40}  & \textbf{31.76} & \textbf{28.54} & \textbf{34.99} & \textbf{32.79} & \textbf{29.78} & \textbf{35.40} & \textbf{32.96} & \textbf{29.89} \\ 
    \bottomrule
    \end{tabular} \\ \addlinespace[0.1cm]
    \begin{tabular}{cc}
     \begin{tabular}{>{\arraybackslash}p{2.1cm}|>{\centering\arraybackslash}p{2.1cm}}
    \toprule
    \textbf{Method} & \textbf{SOTS} \cite{li2018benchmarking} \\
    \midrule
     DehazeNet \cite{cai2016dehazenet} & 22.46\;/\;0.851 \\
     AODNet \cite{li2017aod} & 20.29\;/\;0.877 \\
     FDGAN \cite{dong2020fd} & 23.15\;/\;0.921 \\
     DehazeFormer \cite{song2023vision} & 31.78\;/\;0.977 \\
    \midrule
     Restormer \cite{zamir2022restormer} & 30.87\;/\;0.969 \\
     NAFNet \cite{chen2022simple} & 30.98\;/\;0.970 \\
     FSNet \cite{cui2023image} & 31.11\;/\;0.971 \\
    \midrule
     AirNet \cite{li2022all} & 23.18\;/\;0.900 \\
     PromptIR \cite{potlapalli2023promptir} & 31.31\;/\;0.973 \\
     Perceive-IR \cite{zhang2025perceive} & 31.65\;/\;0.977 \\
     AdaIR \cite{cui2025adair} & 31.80\;/\;0.981 \\
     M\textsuperscript{2}IR &  \textbf{32.12}\;/\;\textbf{0.984} \\
    \bottomrule
    \end{tabular} & \begin{tabular}{>{\arraybackslash}p{2.1cm}|>{\centering\arraybackslash}p{2.1cm}}
    \toprule
    \textbf{Method} & \textbf{Rain100L} \cite{yang2017deep} \\
    \midrule
     UMR \cite{yasarla2019uncertainty} & 32.39\;/\;0.921 \\
     MSPFN \cite{jiang2020multi} & 33.50\;/\;0.948 \\
     LPNet \cite{fu2019lightweight} & 33.61\;/\;0.958 \\
     DRSformer \cite{chen2023learning} & 38.14\;/\;0.983 \\
    \midrule
     Restormer \cite{zamir2022restormer} & 36.74\;/\;0.978 \\
     NAFNet \cite{chen2022simple} & 36.63\;/\;0.977 \\
     FSNet \cite{cui2023image} & 37.27\;/\;0.980 \\
    \midrule
     AirNet \cite{li2022all} & 34.90\;/\;0.977 \\
     PromptIR \cite{potlapalli2023promptir} & 37.04\;/\;0.979 \\
     Perceive-IR \cite{zhang2025perceive} & 38.41\;/\;0.984 \\
     AdaIR \cite{cui2025adair} & 38.90\;/\;0.985 \\
     M\textsuperscript{2}IR &  \textbf{39.33}\;/\;\textbf{0.986} \\
    \bottomrule
    \end{tabular} \\
    \end{tabular} \\
    \end{tabular}
    }
  \label{tab:one_by_one}
\end{table}

Although the proposed method is primarily designed for the all-in-one image restoration task, it also exhibits competitive or even superior performance in single-task scenarios. 
To validate this property, we conduct independent training and comprehensive evaluation on three representative tasks, including denoising, dehazing, and deraining. 
The training and testing datasets are consistent with those used in the all-in-one setting, while the denoising task additionally involves the Urban100 \cite{huang2015single} and Kodak24 \cite{kodak24} datasets for evaluation. 
As presented in Tab.~\ref{tab:one_by_one}, our method consistently achieves the best results across all datasets and noise levels in the denoising task, surpasses state-of-the-art approaches in the dehazing task with a PSNR of 32.12\,dB and an SSIM of 0.984 on the SOTS dataset, and establishes new performance records in the deraining task with 39.33\,dB PSNR and 0.986 SSIM on the Rain100L dataset. 
These results demonstrate that the proposed method maintains strong generalization ability and task adaptability, effectively balancing the capacity for handling diverse degradations with high performance in specialized image restoration scenarios. 

\begin{table}[!htbp]
  \centering
  \caption{Quantitative results on the unseen desnowing task with the CSD\cite{chen2021all} dataset. All models are trained on 3-Degradation settings.}
    \begin{tabular}{c|c|c|c}
    \toprule
    AirNet & PromptIR & AdaIR & M\textsuperscript{2}IR \\
    \midrule
    19.32/0.733 & 20.47/0.764 & 20.54/0.764 & \textbf{20.83}/\textbf{0.770} \\
    \bottomrule
    \end{tabular}
  \label{tab: unseen}
\end{table}

\textbf{Generalization to out-of-distribution degradations.} To verify the generalization of M\textsuperscript{2}IR on unseen tasks, we take the desnowing task as an example and conduct evaluation on the CSD \cite{chen2021all} dataset. 
Tab.~\ref{tab: unseen} shows that M\textsuperscript{2}IR achieves superior performance, demonstrating its strong robustness to out-of-distribution degradations. It further validates the effectiveness of DACP and DGEA in leveraging degradation-aware priors.

\subsection{Ablation Studies}

In this section, we conduct several ablation experiments, which are performed on the image deraining task by training models for 20 epochs using the Rain100L \cite{yang2017deep}. 

\begin{table}[t]
  \centering
  \caption{Ablation Studies for each component.}
    \begin{footnotesize}{
    \begin{tabular}{cccc|cc}
    \toprule
    \multicolumn{1}{c}{\multirow{2}{*}{\textbf{MST}}} & \multicolumn{1}{c}{\multirow{2}{*}{\textbf{DACP}}} & \multicolumn{2}{c|}{\textbf{DGEA}} & \multicolumn{1}{c}{\multirow{2}{*}{\textbf{PSNR}}} & \multicolumn{1}{c}{\multirow{2}{*}{\textbf{SSIM}}} \\
      &   & \multicolumn{1}{c}{\textbf{special}} & \multicolumn{1}{c|}{\textbf{shared}} &   &  \\
    \midrule
     \ding{55} & \ding{55} & \ding{55} & \ding{55} & 38.43 & 0.9823 \\
     \ding{51} & \ding{55} & \ding{55} & \ding{55} & 38.55 & 0.9834 \\
     \ding{51} & \ding{55} & \ding{51} & \ding{55} & 38.67 & 0.9832 \\
     \ding{51} & \ding{55} & \ding{51} & \ding{51} & 38.82 & 0.9833 \\
     \ding{51} & \ding{51} & \ding{51} & \ding{55} & 38.73 & 0.9837 \\
     \ding{55} & \ding{51} & \ding{51} & \ding{51} & 38.71 & 0.9836 \\
     \ding{51} & \textbf{MLP} & \ding{51} & \ding{51} & 38.55 & 0.9825 \\
     \ding{51} & \ding{51} & \ding{51} & \ding{51} & \textbf{38.93} & \textbf{0.9842} \\
    \bottomrule
    \end{tabular}}
    \end{footnotesize}
  \label{tab:ablation_components}
\end{table}

\textbf{Effects of the each component.} As shown in Tab.~\ref{tab:ablation_components}, replacing the Transformer \cite{zamir2022restormer} with MST brings a clear improvement in both PSNR and SSIM, with PSNR increasing from 38.43 to 38.55 and SSIM increasing from 0.9823 to 0.9834, indicating that the selective state modulation proactively suppresses irrelevant information in degraded regions while enhancing the preservation of textures. 
Incorporating the special branch of DGEA yields further gains, increasing PSNR to 38.67 by enabling pixel-wise routing toward degradation-specific experts. 
Introducing the shared branch in addition to the special branch provides complementary global priors, leading to more stable performance across diverse degradations and achieving a PSNR of 38.82. 
When DACP is integrated with the complete DGEA, the routing process becomes guided by more accurate degradation-aware contextual priors, allowing the network to better adapt to spatially variant degradations and reach a PSNR of 38.73 with an SSIM of 0.9837. 
When MST is removed while keeping DACP and both branches of DGEA, the performance drops to 38.71/0.9836, demonstrating that although degradation-aware routing is effective, the selective state modulation mechanism is still indispensable for fully exploiting the routed features and maintaining high-fidelity reconstruction quality.
Moreover, replacing the pre-trained DA-CLIP in DACP with a learnable MLP leads to a significant performance degradation, with PSNR decreasing from 38.93 to 38.55 and SSIM dropping from 0.9842 to 0.9825. This notable decline verifies that the pre-trained degradation-aware representation provided by DA-CLIP is crucial for reliable routing guidance. A randomly initialized MLP fails to capture discriminative degradation semantics, resulting in inferior contextual priors and substantially weakened restoration performance.
The full configuration, which combines MST, DACP, and both branches of DGEA, achieves the highest scores, confirming that these components are mutually reinforcing and together offer the most robust all-in-one restoration performance. 

\begin{table}[t]
  \centering
  \caption{Ablation Studies for the number N of specialized experts in each ADEC module, i.e., excluding the shared expert.}
  \begin{footnotesize}{
    \begin{tabular}{c|ccccc}
    \toprule
    \textbf{Method} & N=2 & N=4 & N=8 & N=16 & N=32 \\
    \midrule
     \textbf{PSNR} & 38.84 & \textbf{38.93} & 38.81 & 38.76 & 38.62 \\
     \textbf{SSIM} & 0.9834 & \textbf{0.9842} & 0.9838 & 0.9836 & 0.9834 \\
    \bottomrule
    \end{tabular}
    }
    \end{footnotesize}
  \label{tab:ablation_expert_number}
\end{table}

\textbf{Effects of the number of experts.} As shown in Tab.~\ref{tab:ablation_expert_number}, the performance first improves as the number of experts increases, reaching the best results when a moderate number of experts is used. 
When the number of experts continues to grow beyond this point, the restoration quality starts to decline slightly. 
This trend suggests that too few experts may limit model capacity, while too many experts could lead to redundancy and less effective expert specialization. 
Considering both performance and computational efficiency, 4 specialized experts provide a balanced trade-off and are therefore adopted in the final model.

\textbf{Effects of the number of the top experts.} Tab.~\ref{tab:ablation_top_k} indicates that selecting an appropriate number of top experts for each input is also important. 
The performance peaks when a moderate number of top experts are activated, while using fewer or more experts per input results in a small but consistent drop in PSNR and SSIM. 
It implies that activating too few experts restricts the diversity of learned representations, whereas activating too many dilutes the expertise of the most relevant specialists. 
The best balance between diversity and specialization is achieved when 2 top experts are selected for each input. 

\begin{table}[t]
  \centering
  \caption{Ablation Studies for the number K of the top experts. }
  \begin{footnotesize}{
    \begin{tabular}{c|cccc}
    \toprule
    \textbf{Method} & K=1 & K=2 & K=3 & K=4 \\
    \midrule
     \textbf{PSNR} & 38.83 & \textbf{38.93} & 38.86 & 38.78 \\
     \textbf{SSIM} & 0.9840 & \textbf{0.9842} & 0.9839 & 0.9837 \\
    \bottomrule
    \end{tabular}
    }
    \end{footnotesize}
  \label{tab:ablation_top_k}
\end{table}

\textbf{Effects of Different Loss Functions.} Tab.~\ref{tab:ablation_losses} presents the results of the ablation study on different loss function configurations. 
When $\ell_{charb}$ is not employed, the default L1 loss is used as the primary reconstruction objective, attaining a PSNR of 38.41 dB and an SSIM of 0.9822, which serves as the baseline. 
Incorporating the $\ell_{charb}$ loss leads to a consistent improvement in both metrics, reaching 38.56 dB in PSNR and 0.9834 in SSIM, confirming its effectiveness in stabilizing optimization and providing more robust pixel-level supervision. 
Adding the $\ell_{FFT}$ loss on top of $\ell_{charb}$ further enhances the results to 38.74 dB and 0.9836, demonstrating that frequency-domain consistency contributes to the preservation of structural detail and the reduction of high-frequency artifacts. 
When the $\ell_{balance}$ loss is combined with $\ell_{charb}$, the performance improves to 38.81 dB and 0.9839, indicating that promoting fair expert activation benefits representation diversity and overall reconstruction quality. 
The best performance is achieved when all three loss components are jointly applied, yielding 38.93 dB in PSNR and 0.9842 in SSIM. 
This suggests that the proposed composite objective effectively integrates pixel-space accuracy, frequency-domain consistency, and balanced expert utilization, leading to superior restoration performance across the evaluation metrics. 

\begin{table}[t]
  \centering
  \caption{Ablation Studies for each loss function. }
  \begin{footnotesize}{
    \begin{tabular}{ccc|cc}
    \toprule
    $\boldsymbol{\ell_{charb}}$ & $\boldsymbol{\ell_{balance}}$ & $\boldsymbol{\ell_{FFT}}$ & \textbf{PSNR} & \textbf{SSIM} \\
    \midrule
    \ding{55} & \ding{55} & \ding{55} & 38.41 & 0.9822 \\
    \ding{51} & \ding{55} & \ding{55} & 38.56 & 0.9834 \\
    \ding{51} & \ding{55} & \ding{51} & 38.74 & 0.9836 \\
    \ding{51} & \ding{51} & \ding{55} & 38.81 & 0.9839 \\
    \ding{51} & \ding{51} & \ding{51} & \textbf{38.93} & \textbf{0.9842} \\
    \bottomrule
    \end{tabular}
    }
    \end{footnotesize}
  \label{tab:ablation_losses}
\end{table}

\begin{table}[t]
  \centering
  \caption{Ablation Studies for the weight $\lambda_1$.}
  \begin{footnotesize}{
    \begin{tabular}{c|ccccc}
    \toprule
    \textbf{Method} & 0.001 & 0.005 & 0.01 & 0.05 & 0.1 \\
    \midrule
     \textbf{PSNR} & 38.89 & 38.74 & \textbf{38.93} & 38.46 & 38.60 \\
     \textbf{SSIM} & 0.9841 & 0.9838 & \textbf{0.9842} & 0.9823 & 0.9830 \\
    \bottomrule
    \end{tabular}
    }
    \end{footnotesize}
  \label{tab:ablation_weight_balance}
\end{table}

\begin{table}[t]
  \centering
  \caption{Ablation Studies for the weight $\lambda_2$.}
  \begin{footnotesize}{
    \begin{tabular}{c|ccccc}
    \toprule
    \textbf{Method} & 0.01 & 0.05 & 0.1 & 0.5 & 1 \\
    \midrule
     \textbf{PSNR} & 38.63 & 38.90 & \textbf{38.93} & 38.71 & 38.72 \\
     \textbf{SSIM} & 0.9835 & 0.9841 & \textbf{0.9842} & 0.9837 & 0.9830 \\
    \bottomrule
    \end{tabular}
    }
    \end{footnotesize}
  \label{tab:ablation_weight_fft}
\end{table}

\textbf{Effects of Different Loss Function Weights.} To investigate the influence of the auxiliary loss weights, we conducted a series of exploratory experiments. 
The ablation results for the load-balancing weight $\lambda_1$ are presented in Tab.~\ref{tab:ablation_weight_balance}, and the results for the frequency-domain weight $\lambda_2$ are shown in Tab.~\ref{tab:ablation_weight_fft}. 
In Tab.~\ref{tab:ablation_weight_balance}, $\lambda_1$ achieves its best performance at 0.01, where the model reaches 38.93 dB PSNR and 0.9842 SSIM. Smaller or larger values lead to slight performance drops, which suggests that very small weights cannot effectively correct expert imbalance, while overly large weights impose excessive regularization on the routing. 
In Tab.~\ref{tab:ablation_weight_fft}, $\lambda_2$ obtains the highest PSNR and SSIM at 0.1, and deviations from this value cause performance degradation, indicating that a moderate emphasis on frequency consistency is most beneficial. 

\subsection{Effectiveness Analysis}

\begin{figure}[t]
    \centering
    \includegraphics[width=1\linewidth]{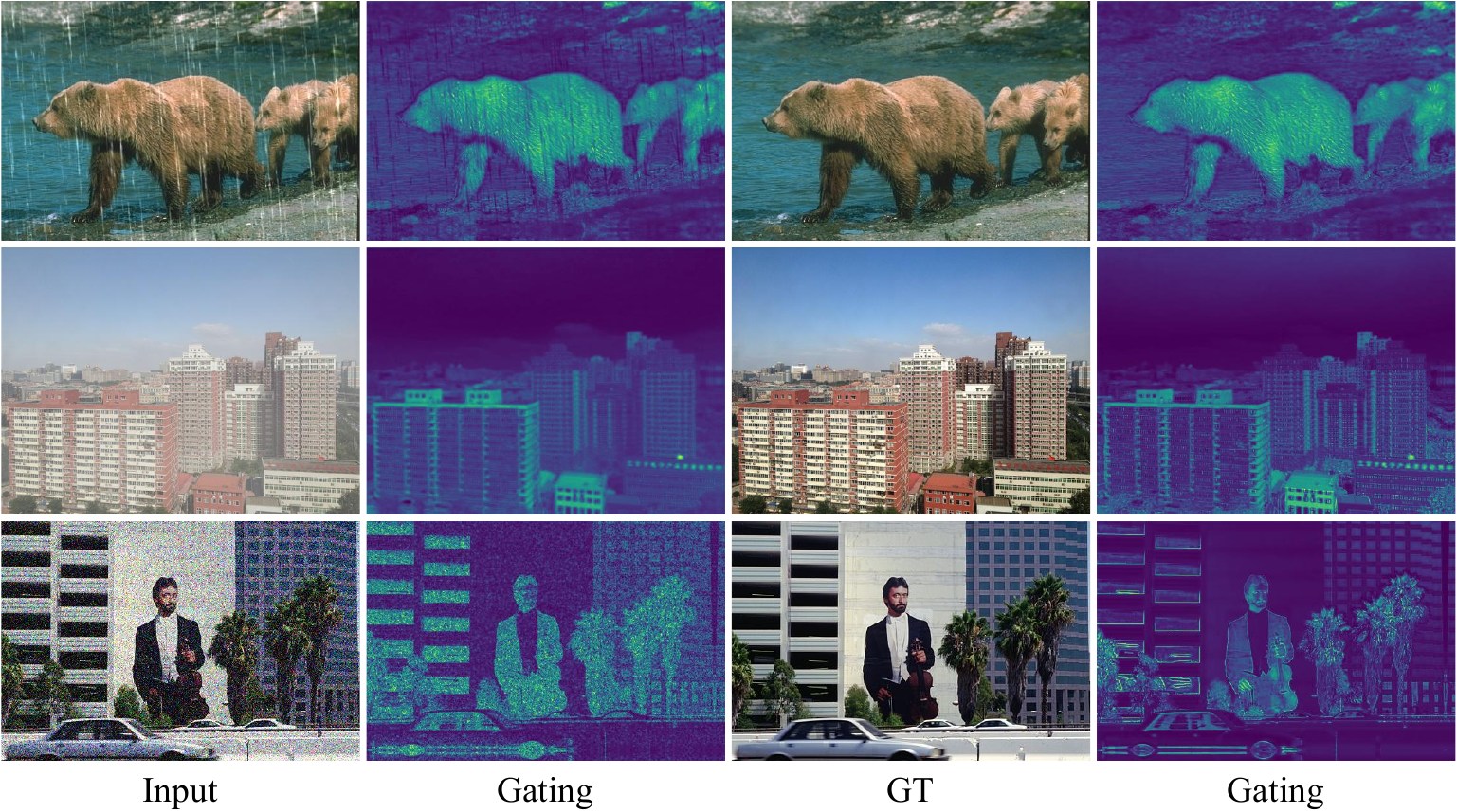}
    \caption{Gating visualization of degraded input images and their corresponding GT during the encoding stage, illustrating the selective state modulation mechanism in MST. The model is trained on three degradation types.}
    \label{fig: gate_effectiveness}
\end{figure}

As illustrated in Fig.~\ref{fig: gate_effectiveness}, during the encoding stage, the gating mechanism exhibits distinct response patterns for different inputs. 
When presented with ground truth (GT) images, the gating maps demonstrate high response values for clean scene content, indicating their capability to accurately identify and preserve structural information. 
Conversely, when processing degraded images, the gating maps show low responses in degraded regions while maintaining high activation on clean scene content. 
This contrasting behavior validates that MST effectively leverages its selective state modulation mechanism, where the gating mechanism serves as its concrete implementation, to proactively suppress the encoding and propagation of degradation-related signals during the encoding phase. 

\begin{figure*}[t]
    \centering
    \subfloat[The encoder composed of Transformer \cite{zamir2022restormer} blocks, exemplified here by AdaIR, which employs this encoder type.]{
        \includegraphics[width=0.4\linewidth]{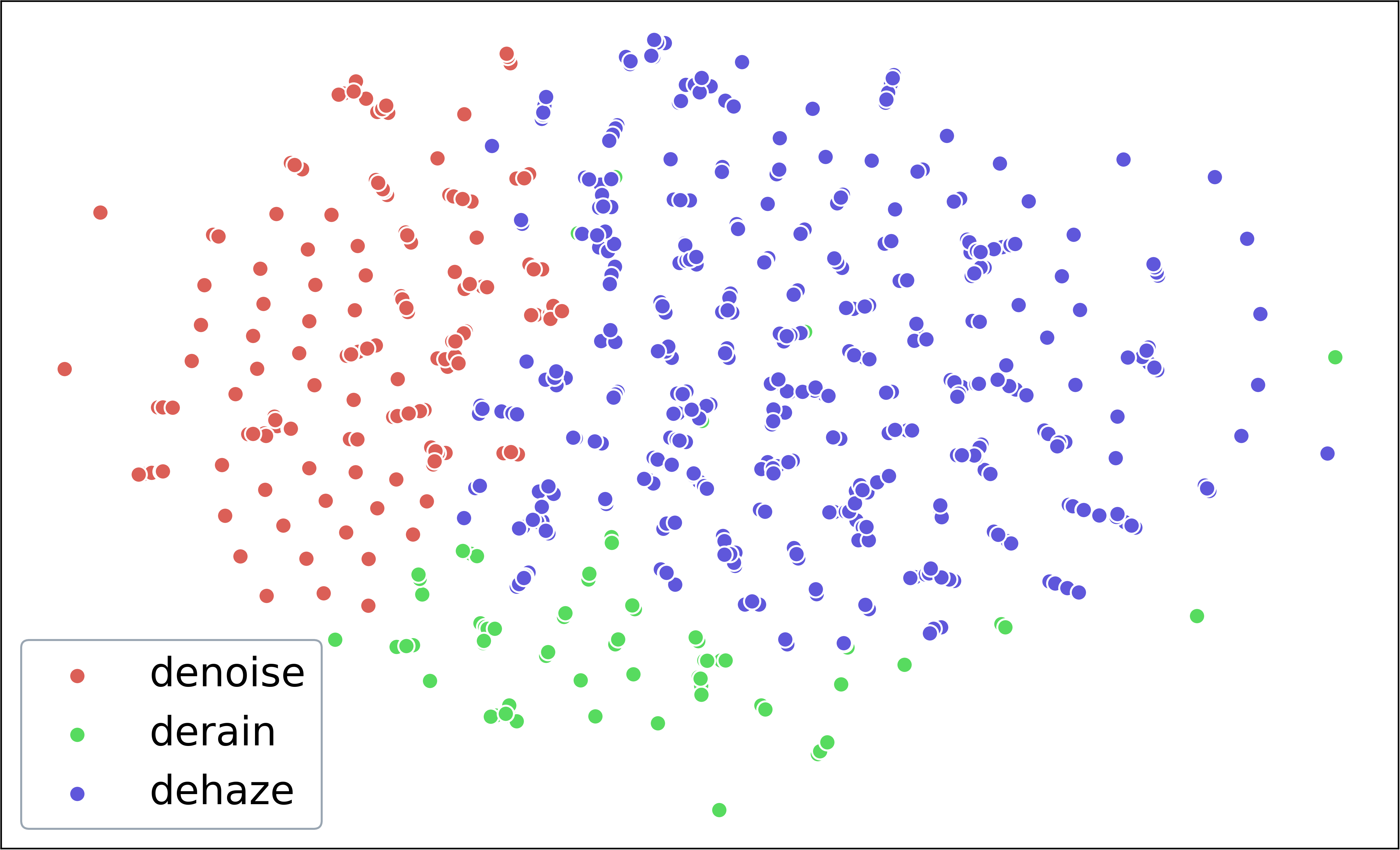}
        \label{fig:sub1}
    }
    \hfil
    \subfloat[The encoder composed of Mamba-Style Transformer blocks.]{
        \includegraphics[width=0.4\linewidth]{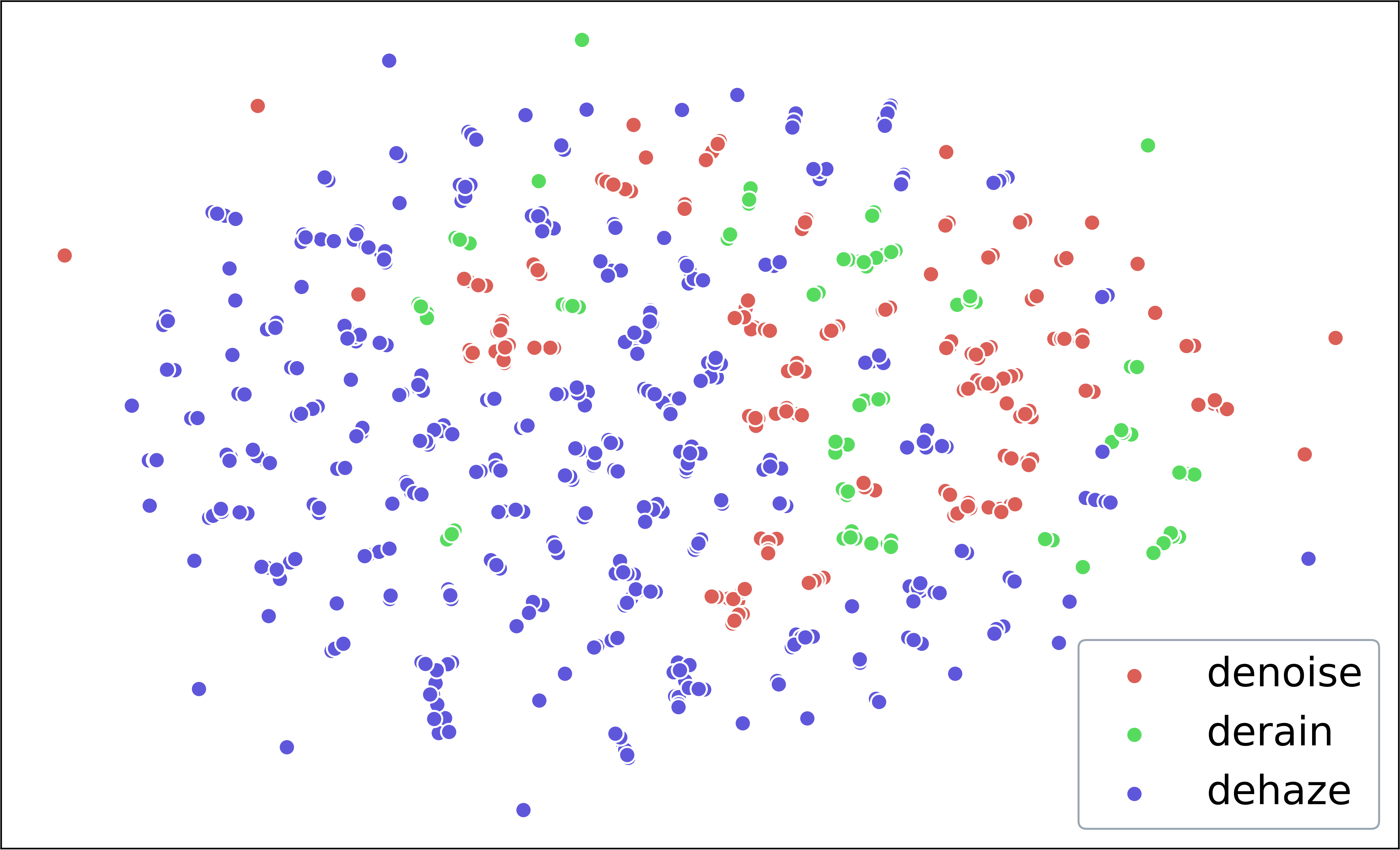}
        \label{fig:sub2}
    }
    \caption{The t-SNE visualisations of the two encoders under three degeneration types.}
    \label{fig: tsne}
\end{figure*}

The t-SNE analysis in Fig.~\ref{fig: tsne} further provides the clear evidence of the effectiveness of our MST. 
When using the Transformer-based \cite{zamir2022restormer} encoder, the features corresponding to denoise, derain, and dehaze tasks form well-separated clusters, indicating that degradation-specific patterns remain highly distinguishable in the latent space. It suggests that this kind of Transformer tends to preserve and propagate degradation, confirming their inherently reactive nature. 
In contrast, the MST-based encoder exhibits highly overlapping and entangled distributions across all degradation types, demonstrating that our selective state modulation can proactively and successfully suppress degradation-related patterns during encoding instead of allowing them to dominate feature learning. 
This entanglement reflects that MST shifts the representation toward structural and content-centric information rather than degradation patterns, thereby producing cleaner, more degradation-agnostic latent representations. 
Such proactive suppression not only validates the core design principle of MST but also reduces the burden on the decoder, allowing ADEC to focus solely on refining residual degradations and ultimately contributing to the superior generalization and restoration performance of M\textsuperscript{2}IR across diverse degradation scenarios. 

\subsection{Efficiency Analysis}

\begin{table}[t]
  \centering
  \caption{Efficiency analysis of model parameters, FLOPs, inference latency, and peak memory usage on both CPU and GPU, evaluated using $720 \times 480$ inputs on an NVIDIA RTX 4090 GPU.}
  \resizebox{\columnwidth}{!}{
    \begin{tabular}{cccccc}
    \toprule
    Method & Params. & FLOPs & latency & CPU & GPU \\
    \midrule
    IDR & 42.3M & 1522G & 384ms & 1387M & 6952M \\
    PromptIR & 34.1M & 745G & 282ms & 1242M & 3127M \\
    GridFormer & 34.1M & 1941G & 746ms & 1355M & 3830M \\
    Pool-AIO & 26.1M & 745G & 293ms & 1246M & 3096M \\
    AdaIR & 28.8M & 778G & 333ms & 1296M & 3110M \\
    MoCE-IR & 25.4M & 474G & 263ms & 1312M & 1253M \\
    MEASNet & 31.7M & 892G & 349ms & 1699M & 3228M \\
    Perceive-IR & 181.2M & 1188G & 399ms & 1996M & 4194M \\
    Perceive-IR (w/o CLIP) & 30.0M & 1176G & 387ms & 2015M & 3607M \\
    \midrule
    M\textsuperscript{2}IR & 285.3M & 975G & 339ms & 3262M & 4103M \\
    M\textsuperscript{2}IR (w/o DA-CLIP) & 39.1M & 961G & 330ms & 1607M & 3149M \\
    \bottomrule
    \end{tabular}
    }
  \label{tab: efficiency}
\end{table}

As presented in Tab.~\ref{tab: efficiency}, we report the model parameters and measure the computational overhead. 
All experiments are conducted on an NVIDIA RTX 4090 GPU with input resolution of $720 \times 480$. 
Although the integration of DA-CLIP introduces a substantial increase in model parameters from 39.1M to 285.3M, the additional computational cost remains minimal. 
Specifically, FLOPs only increase marginally from 961G to 975G, and inference latency increases merely by 9ms from 330ms to 339ms. 
This demonstrates that our degradation-aware contextual priors effectively leverage pre-trained knowledge without imposing significant computational burdens during inference, making M\textsuperscript{2}IR a practical solution for real-world all-in-one image restoration applications. 

%% file: sec/5_conclusion.tex
\section{Conclusion}
In this paper, we propose M\textsuperscript{2}IR, a novel all-in-one image restoration framework that redefines the traditional passive compensation paradigm as a proactive degradation regulation paradigm. 
At its core, the Mamba-Style Transformer (MST) employs selective state modulation with input-dependent dynamic weighting, effectively suppressing degradation-induced interference while maintaining stable and reliable feature transmission. 
To further enhance adaptability across diverse degradation scenarios, we introduce the Adaptive Degradation Expert Collaboration (ADEC) module, which integrates domain-specific experts with a shared expert. 
By dynamically activating the most relevant domain-specific expert and shared experts, ADEC facilitates task-aware decoding tailored to distinct degradation types. Extensive experiments on multiple all-in-one benchmarks demonstrate that M\textsuperscript{2}IR achieves new state-of-the-art performance, delivering superior detail preservation and faithful recovery under spatially varying degradations.